\documentclass[conference]{IEEEtran}
\input{glyphtounicode}
\usepackage{times}

\usepackage[numbers]{natbib}
\usepackage{multicol}
\usepackage{graphicx}
\usepackage{amsmath}
\usepackage{amssymb}
\usepackage{xcolor}
\usepackage{tikz}
\usepackage{array}
\usepackage{multirow}
\usepackage{booktabs}
\usepackage{xspace}
\usepackage{microtype}
\usepackage{subcaption}
\usepackage[bookmarks=true,pdfencoding=auto,breaklinks,colorlinks,hypertexnames=false]{hyperref}
\usepackage{cleveref}
\crefname{section}{Sec.}{Secs.}
\Crefname{section}{Section}{Sections}
\Crefname{table}{Table}{Tables}
\crefname{table}{Tab.}{Tabs.}
\Crefname{figure}{Fig}{Figures}
\crefname{figure}{Fig.}{Figs.}

\hypersetup{
   pdfauthor={Rohit Mohan, Florian Drews, Yakov Miron, Daniele Cattaneo, Abhinav Valada},
   pdftitle={UP-Fuse: Uncertainty-guided LiDAR-Camera Fusion for 3D Panoptic Segmentation},
   pdfsubject={3D Panoptic Segmentation},
   pdfkeywords={Multimodal learning; Computer Vision for Robotics; Autonomous driving}
}

\begin{document}

\title{UP-Fuse: Uncertainty-guided LiDAR-Camera Fusion for 3D Panoptic Segmentation}

\author{\authorblockN{Rohit Mohan\authorrefmark{1},
Florian Drews\authorrefmark{2},
Yakov Miron\authorrefmark{2}\authorrefmark{3}, 
Daniele Cattaneo\authorrefmark{1} and
Abhinav Valada\authorrefmark{1}}
\authorblockA{\authorrefmark{1}
University of Freiburg,
Germany\\ Email: \{mohan,cattaneo,valada\}@cs.uni-freiburg.de}
\authorblockA{\authorrefmark{2}Robert Bosch GmbH, Germany\\
Email: \{florian.drews,yakov.miron\}@de.bosch.com}
\authorblockA{\authorrefmark{3}University of Haifa}}

\maketitle

\begin{abstract}
LiDAR-camera fusion enhances 3D panoptic segmentation by leveraging camera images to complement sparse LiDAR scans, but it also introduces a critical failure mode. Under adverse conditions, degradation or failure of the camera sensor can significantly compromise the reliability of the perception system.
To address this problem, we introduce UP-Fuse, a novel uncertainty-aware fusion framework in the 2D range-view that remains robust under camera sensor degradation, calibration drift, and sensor failure. Raw LiDAR data is first projected into the range-view and encoded by a LiDAR encoder, while camera features are simultaneously extracted and projected into the same shared space. At its core, UP-Fuse employs an uncertainty-guided fusion module that dynamically modulates cross-modal interaction using predicted uncertainty maps.
These maps are learned by quantifying representational divergence under diverse visual degradations, ensuring that only reliable visual cues influence the fused representation.
The fused range-view features are decoded by a novel hybrid 2D-3D transformer that mitigates spatial ambiguities inherent to the 2D projection and directly predicts 3D panoptic segmentation masks.
Extensive experiments on Panoptic nuScenes, SemanticKITTI, and our introduced Panoptic Waymo benchmark demonstrate the efficacy and robustness of UP-Fuse, which maintains strong performance even under severe visual corruption or misalignment, making it well suited for robotic perception in safety-critical settings.
We make the code and models publicly available at \url{http://upfuse.cs.uni-freiburg.de}.\@
\end{abstract}

\IEEEpeerreviewmaketitle%

\section{Introduction}\label{sec:intro}

\begin{figure}
  \centering
  \includegraphics[width=\linewidth]{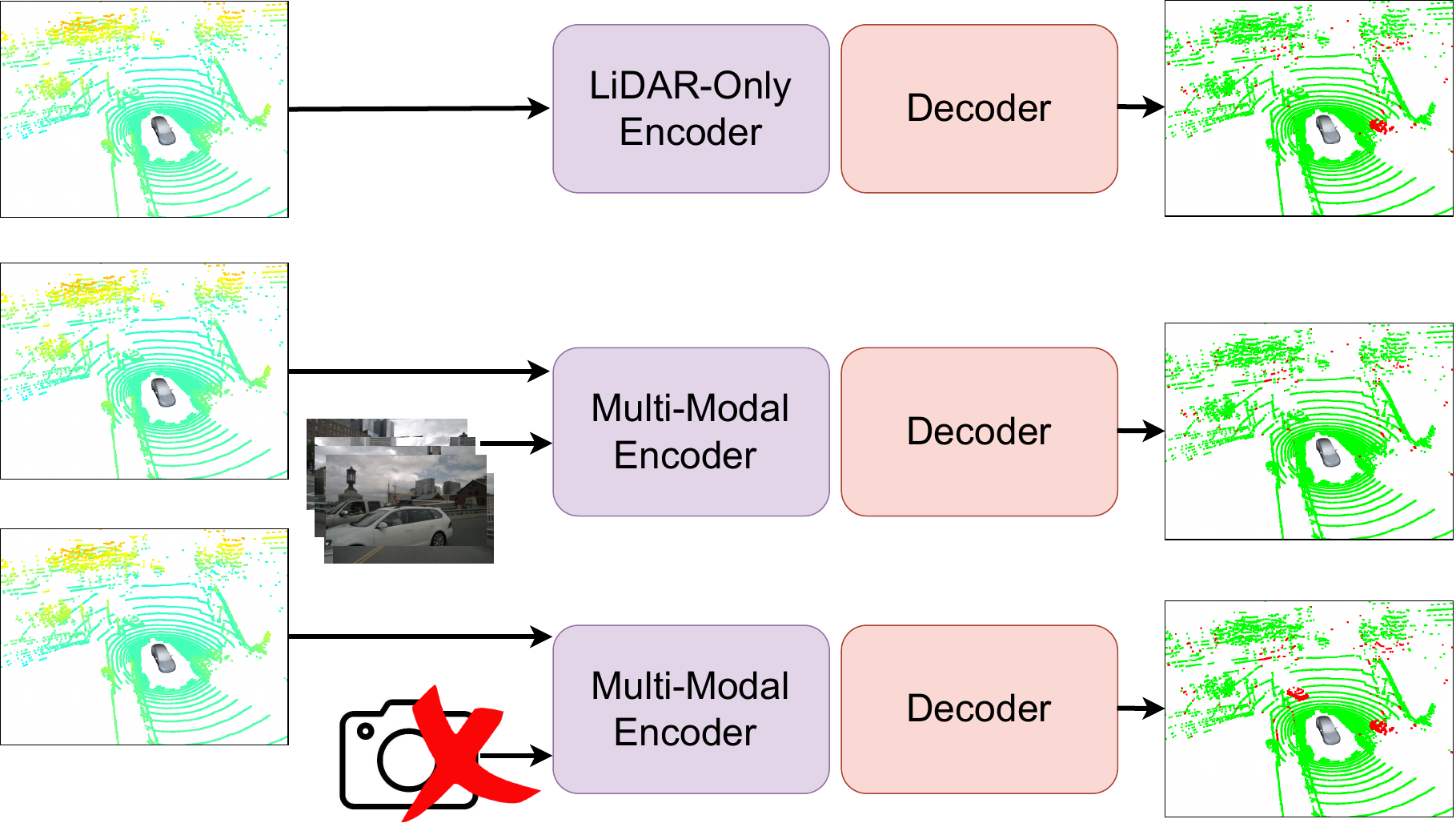} 
  \caption{Visualization of 3D panoptic segmentation: green indicates correct and red denotes errors. LiDAR-camera fusion significantly enhances segmentation over LiDAR-only methods, accurately detecting previously missed vehicles. However, camera sensor failure scenarios reveal a critical vulnerability, with fusion-based performance falling below LiDAR-only baselines and failing to detect previously identified objects. This highlights the crucial need for both relevance and reliability in multi-modal perception.}\label{fig:teaser}

\end{figure}

3D panoptic segmentation~\cite{kirillov2019panoptic, milioto2020lidar} unifies semantic and instance understanding of complex scenes, making it central for robotic perception, including autonomous driving~\cite{neshev20253d}. LiDAR offers precise geometric measurements, but its sparsity and lack of appearance cues hinder reliable segmentation of small, distant, or geometrically similar objects. Incorporating dense, high-resolution camera data can mitigate these limitations by providing complementary texture and color information~\cite{valada2016convoluted, schramm2024bevcar, valada2016towards}. However, achieving effective fusion remains challenging, as the model has to learn not only \textit{where} and \textit{what} information to fuse, but also \textit{when} to trust each modality. This reliability awareness is crucial under adverse conditions such as sensor failure, calibration errors, or visual domain shift, where camera inputs may become unreliable, as illustrated in \cref{fig:teaser}.

Existing fusion approaches for 3D panoptic segmentation are ill-equipped to handle such scenarios and thus degrade sharply when a modality fails. A crucial gap lies in developing a fusion paradigm that can discern not only which features are relevant but also whether they are valid, enabling adaptive and context-dependent integration across modalities.
To address these challenges, we propose UP-Fuse (\textbf{U}ncertainty-aware \textbf{P}anoptic \textbf{Fus}ion), an uncertainty-aware multi-modal fusion framework that leverages the range-view projection space. Our core contribution is an Uncertainty-Aware Fusion Module that jointly learns to evaluate cross-modal relevance and visual reliability. The module integrates deformable attention to identify informative visual features with an uncertainty head that estimates feature-level confidence. The fusion module is trained to associate feature patterns induced by camera sensor degradations (e.g., dropouts, occlusions, or out-of-domain distortions) with their impact on representational fidelity.
The predicted uncertainty dynamically modulates the contribution of image features, enabling the network to attenuate unreliable cues while retaining informative ones. The fused representation is then processed by a Hybrid 2D-3D Panoptic Decoder that directly predicts 3D panoptic masks while alleviating projection ambiguities and boundary discontinuities inherent to the 360° range-view representation. 

We extensively evaluate UP-Fuse on the Panoptic nuScenes, SemanticKITTI, and Waymo Open Dataset~\cite{sun2020scalability} benchmarks. 
Since Waymo does not provide panoptic annotations, we generate them using the publicly available semantic segmentation and 3D bounding box labels, and establish Panoptic Waymo, a new multi-modal 3D panoptic benchmark with several baselines. 
The main contributions of this work are summarized as follows: (1) the UP-Fuse framework for uncertainty-aware multi-modal 3D panoptic segmentation, (2) an uncertainty-guided fusion module that enables reliability-aware integration of LiDAR and camera features, (3) a hybrid 2D-3D panoptic decoder that alleviates projection ambiguities in the 360° range-view representation, (4) a new multi-modal 3D panoptic benchmark for the Waymo Open Dataset with derived annotations and strong baselines, and (5) publicly released code and models upon acceptance.

\section{Related Work}\label{sec:related_work}

{\parskip=2pt
\noindent\textbf{LiDAR Panoptic Segmentation}: 
LiDAR Panoptic segmentation methods can be broadly categorized as top-down, bottom-up, or transformer-based approaches. Top-down methods~\cite{sirohi2021efficientlps, xu2023aop, ye2023lidarmultinet} typically decompose the task into separate sub-tasks, using dedicated heads for instance and semantic segmentation. The instance head predicts object-level representations such as 2D instance masks from range-view (RV) projections~\cite{sirohi2021efficientlps} or 3D bounding boxes and instance masks in voxelized space~\cite{xu2023aop, ye2023lidarmultinet}, while the semantic head produces dense per-point or per-pixel class predictions. The outputs from both heads are then fused through heuristic-based modules~\cite{kirillov2019panoptic} to generate the final panoptic output. In contrast, bottom-up methods~\cite{hong2021lidar, li2022panoptic, razani2021gp, xu2022sparse, zhou2021panoptic} jointly predict semantic labels and instance-related cues for all points within a single network. They typically regress offsets from each \textit{thing} point toward its object center~\cite{hong2021lidar, li2022panoptic, mohan2025open, zhou2021panoptic} or learn affinities between points~\cite{razani2021gp}, followed by clustering or grouping to form complete instances~\cite{mohan2022perceiving}.}

Transformer-based approaches~\cite{gu2022maskrange, marcuzzi2023mask, xiao2025position, su2023pups} unify semantic and instance segmentation through a shared mask-classification framework, using learned queries to predict panoptic masks in an end-to-end manner. These methods differ in input representation, including point-based~\cite{su2023pups}, sparse voxel~\cite{marcuzzi2023mask}, and RV projections~\cite{gu2022maskrange}. The latter offers computational efficiency by transforming sparse 3D points into dense 2D grids, enabling the use of mature 2D segmentation architectures. However, lifting 2D predictions back into 3D heuristically, often introduces artifacts and struggles with occlusion~\cite{sirohi2021efficientlps, gu2022maskrange}. Our approach follows the RV transformer paradigm but incorporates a novel 2D-3D hybrid panoptic decoder, allowing more accurate and context-aware 3D panoptic segmentation. 

{\parskip=2pt
\noindent\textbf{Multi-Modal LiDAR Panoptic Segmentation}: 
Panoptic-FusionNet~\cite{song2024panoptic} integrates camera features into a 3D backbone through multi-scale point-voxel-pixel correspondence tables, demonstrating the benefit of incorporating visual cues into LiDAR-based panoptic segmentation.
Building on this idea, LCPS~\cite{zhang2023lidar} introduces a voxel-space fusion pipeline that combines asynchronous temporal compensation, semantic-aware region alignment, and point-to-voxel propagation to enable content-aware cross-modal fusion.
More recently, IAL~\cite{pan2025images} proposes a geometry-aware fusion architecture that performs cross-modal interaction through token-level attention between LiDAR and image representations. By incorporating modality-synchronized data augmentation during training, IAL encourages stronger geometric alignment between modalities. While these methods demonstrate the value of multi-modal fusion, Panoptic-FusionNet relies on static geometric associations~\cite{song2024panoptic} that fuse features without considering their contextual relevance. Both LCPS~\cite{zhang2023lidar} and IAL~\cite{pan2025images} emphasize relevance-based fusion through content-aware feature selection or geometry-aware token interaction but lack an explicit mechanism to assess the reliability of visual features under adverse conditions. Camera feature reliability can be affected by sensor imperfections, adverse lighting, and environmental conditions, which introduce aleatoric noise~\cite{kendall2017uncertainties}. Existing fusion methods do not explicitly account for such variations in feature reliability, while traditional uncertainty estimation techniques such as Bayesian neural networks~\cite{blundell2015weight} or deep ensembles~\cite{lakshminarayanan2017simple} are computationally expensive and primarily target epistemic uncertainty. In our work, we model uncertainty at the feature level, enabling the network to quantify the reliability of camera-encoded features. UP-Fuse introduces an uncertainty-guided fusion module that integrates both relevance- and reliability-based fusion. By coupling cross-modal interaction with uncertainty awareness, our method adaptively balances the contributions of each modality. This design aligns with recent efforts to evaluate robustness under sensor degradation and failure~\cite{mohan2024progressive, sodano2023robust, yan2023cross}.}

\section{UP-Fuse Architecture}
\begin{figure*}[h]
  \centering
  \includegraphics[width=\linewidth]{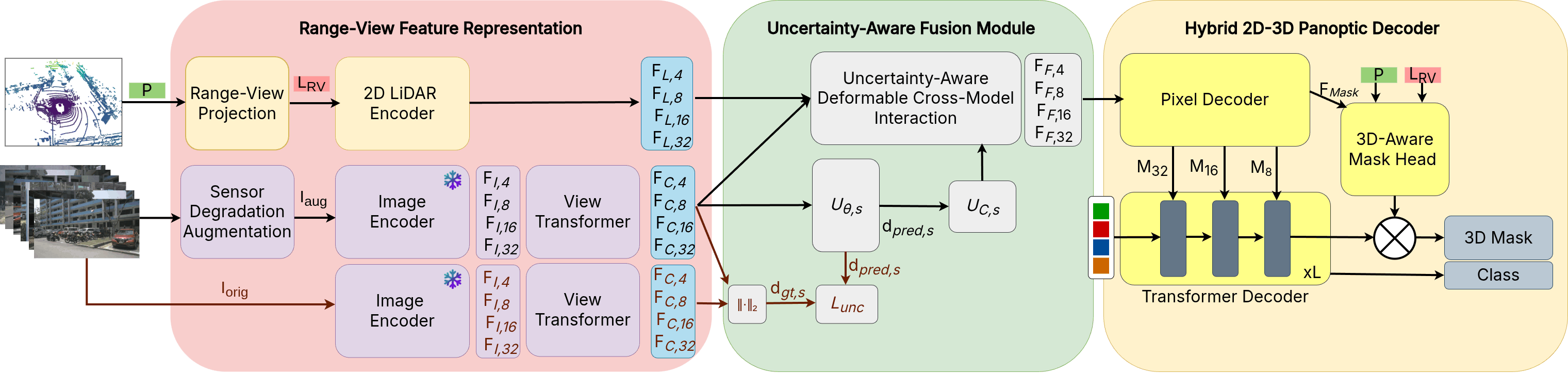}
  \caption{Illustration of the proposed \textbf{UP-Fuse} architecture. LiDAR and multi-view camera images are fused onto a shared space of range-view feature representations. The Uncertainty-Aware Fusion Module adaptively integrates modalities via uncertainty-weighted deformable cross-modal interaction to attenuate unreliable visual cues. Finally, a Hybrid 2D-3D Panoptic Decoder generates 3D predictions. Paths and blocks shown in \textcolor{brown}{brown} are used only during training.}\label{fig:arch}

\end{figure*}

Our proposed UP-Fuse architecture is illustrated in \cref{fig:arch}. The network first projects both LiDAR and camera data into a unified 2D range-view (RV) representation, enabling dense, pixel-aligned feature extraction (\cref{sec:lidar_projection_encoding} and \cref{sec:image_encoding_transformation}). These multi-scale, aligned features are then processed by our Uncertainty-Aware Fusion Module, which adaptively combines the modalities based on both cross-modal relevance and visual reliability (\cref{sec:fusion}). Finally, the resulting fused features are fed into our novel Hybrid 2D-3D Panoptic Decoder. This decoder generates direct 3D panoptic predictions for the original point cloud, while addressing projection ambiguities and 360° wrap-around discontinuities (\cref{sec:decoder}). We detail these components in the following sections.

\subsection{Range-View Feature Representation}\label{sec:view_representation}

\subsubsection{LiDAR Range-View Projection and Encoding}\label{sec:lidar_projection_encoding}

{\parskip=2pt
\textbf{Range-View Projection}:
The raw LiDAR point cloud $\mathbf{P} \in \mathbb{R}^{N \times 4}$ consists of $N$ points, where each point $\mathbf{p}_j = (x_j, y_j, z_j, i_j)$ contains the Cartesian coordinates $(x_j, y_j, z_j)$ and the intensity value $i_j$ of the reflected laser beam. The 3D point cloud is projected into a dense 2D spherical representation, referred to as the range-view image $\mathbf{L}_{\text{RV}}$, with spatial resolution $H \times W$. Each 3D point $\mathbf{p}_j$ is mapped to pixel coordinates $(u_j, v_j)$ by computing its horizontal azimuth angle $\theta_j$ and vertical elevation angle $\phi_j$. The range is defined as $r_j = \sqrt{x_j^2 + y_j^2 + z_j^2}$, and the angular components are given by:}
\begin{equation}
    \theta_j = -\arctan(y_j, x_j), \quad \phi_j = \arcsin\left(\frac{z_j}{r_j}\right).
\label{eq:rv1}
\end{equation}
The angles are normalized according to the LiDAR sensor’s field of view. 
Here, $f_{\text{up}}$ and $f_{\text{down}}$ denote the vertical angular limits above and below the horizontal plane, 
while $f_{\text{left}}$ and $f_{\text{right}}$ represent the horizontal limits to the left and right of the forward axis. 
The total spans are $f_h = |f_{\text{left}}| + |f_{\text{right}}|$ and $f_v = |f_{\text{up}}| + |f_{\text{down}}|$, 
which are used to scale the angles to the image dimensions:

\begin{equation}
\begin{bmatrix}
    u_j \\ v_j
\end{bmatrix}
=
\begin{bmatrix}
    \left( \frac{\theta_j + |f_{\text{left}}|}{f_h} \right) W \\
    \left( 1 - \frac{\phi_j + |f_{\text{down}}|}{f_v} \right) H
\end{bmatrix}.
\label{eq:rv2}
\end{equation}
The resulting range-view image $\mathbf{L}_{\text{RV}}$ is constructed by assigning to each pixel $(u_j, v_j)$ the corresponding point features~\cite{hindel2025label}:
\begin{equation}
\mathbf{L}_{\text{RV}}(u_j, v_j) = [r_j, z_j, i_j].
\end{equation}
To handle occlusions where multiple points map to the same pixel, all points are sorted by decreasing range and projected sequentially, ensuring that only the nearest point (with minimum $r_j$) is retained to maintain visibility consistency. This projection defines a mapping $\mathcal{P}_{\text{3D} \rightarrow \text{RV}}: \mathbf{p}_j \rightarrow (u_j, v_j)$, and we store its inverse $\mathcal{P}_{\text{RV} \rightarrow \text{3D}}$, which is later used in the 2D-3D hybrid decoder (see \cref{sec:decoder}) to lift 2D features back into 3D space.

{\parskip=2pt
\noindent\textbf{LiDAR Feature Encoding}:
The projected range-view image $\mathbf{L}_{\text{RV}}$ is processed by a 2D LiDAR encoder to extract hierarchical feature representations. We employ a Swin Transformer~\cite{liu2021swin} backbone that effectively captures both local geometric structures and global spatial dependencies. The encoder produces feature maps at four resolution levels, denoted as $\mathbf{F}_{L,4}$, $\mathbf{F}_{L,8}$, $\mathbf{F}_{L,16}$, and $\mathbf{F}_{L,32}$, corresponding to output strides of 4, 8, 16, and 32 relative to the input resolution. These multi-scale features are subsequently fed to the Uncertainty-Aware Fusion Module (see \cref{sec:fusion}).}

\subsubsection{Image Encoding and View Transformation}\label{sec:image_encoding_transformation}

{\parskip=2pt
\noindent\textbf{Image Encoding}: 
Each of the $M$ calibrated camera views with spatial dimensions $C_H \times C_W$ is independently processed by an image encoder to capture rich visual cues. We adopt a Swin Transformer~\cite{liu2021swin} backbone that outputs multi-scale feature maps at four hierarchical stages, denoted as $\mathbf{F}_{I,4}$, $\mathbf{F}_{I,8}$, $\mathbf{F}_{I,16}$, and $\mathbf{F}_{I,32}$, corresponding to feature strides of 4, 8, 16, and 32 relative to the original image resolution. The backbone is pre-trained and kept frozen during training. The extracted features from all camera views are then passed to the view transformation module for geometric alignment with the LiDAR range-view representation.}

{\parskip=2pt
\noindent\textbf{View Transformation}:
To enable spatial alignment between image and LiDAR representations, we establish a dense correspondence between the multi-view image planes and the LiDAR range-view image. This is accomplished by constructing a pseudo point cloud from the camera views and then projecting it into the LiDAR range-view space using the same mapping described in \cref{sec:lidar_projection_encoding}. Specifically, the LiDAR point cloud $\mathbf{P}$ is first projected into each of the $M$ camera views to generate $M$ sparse depth maps $\mathcal{D}_{\text{sparse}}$. These are densified using the depth completion method of~\cite{ku2018defense}, resulting in dense depth maps $\mathcal{D}_{\text{dense}}$. Each pixel $(i,j)$ in view $m$ is then back-projected into a 3D point expressed in the LiDAR coordinate frame:}
\begin{equation}
\mathbf{p} = \mathbf{T}_{m}^{-1} 
\begin{bmatrix}
    \mathcal{D}_{\text{dense}}(i,j) \cdot \mathbf{I}_{m}^{-1} \cdot {\left[i, j, 1\right]}^{T} \\
    1
\end{bmatrix},
\end{equation}
where $\mathbf{I}_{m}$ is the $3 \times 3$ intrinsic matrix of camera $m$, and $\mathbf{T}_{m}$ is the $4 \times 4$ extrinsic matrix transforming from the camera to LiDAR frame. This produces a dense, camera-derived point cloud $\mathbf{P}_{\text{cam}}$ that aggregates information from all $M$ views. We then project $\mathbf{P}_{\text{cam}}$ into the $H \times W$ range-view using the same projection formulation as in \cref{eq:rv1} and \cref{eq:rv2}, yielding a dense mapping $\mathcal{M}: (m, i, j) \rightarrow (u, v)$ from each camera pixel to a corresponding range-view pixel. This mapping enables warping of the multi-scale image features $\mathbf{F}_{I,s}$ into the range-view domain to obtain RV-aligned image features $\mathbf{F}_{C,s}$ at each scale $s \in \{4, 8, 16, 32\}$. If multiple pixels project to the same RV location, their features are averaged. The aligned features serve as multi-modal counterparts to the LiDAR features $\mathbf{F}_{L,s}$ and are forwarded to the Uncertainty-Aware Fusion Module (see \cref{sec:fusion}).

\begin{figure}[t]
\centering
\setlength{\tabcolsep}{2pt}
\renewcommand{\arraystretch}{1.0}

\begin{tabular}{@{}m{2.8cm} m{4.7cm}@{}}
\centering\scriptsize \textbf{Input Camera Views} & \centering\scriptsize \textbf{Predicted Uncertainty Heatmap} \\
\end{tabular}

\begin{tabular}{@{}m{2.8cm} m{4.7cm}@{}}
\includegraphics[width=\linewidth]{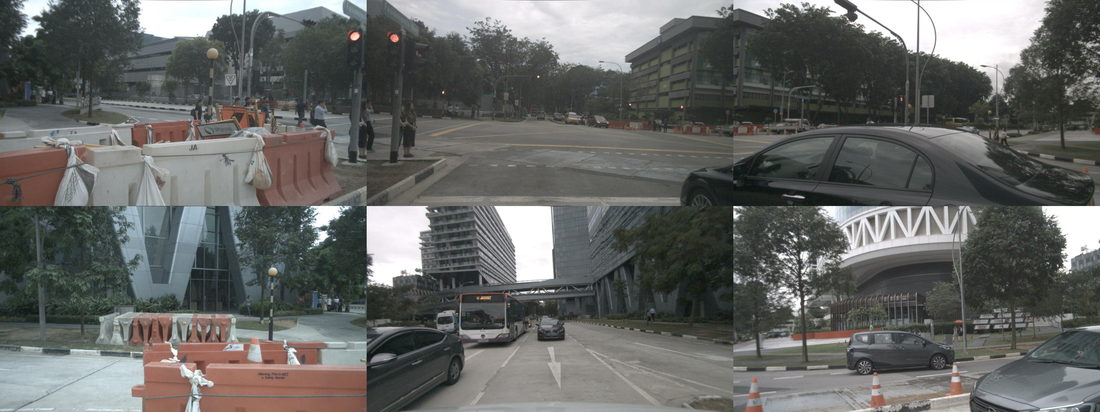} &
\includegraphics[width=\linewidth]{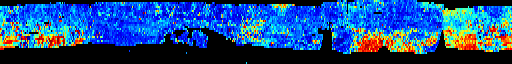} \\
\end{tabular}

\scriptsize (a) Original \\[2pt]

\begin{tabular}{@{}m{2.8cm} m{4.7cm}@{}}
\includegraphics[width=\linewidth]{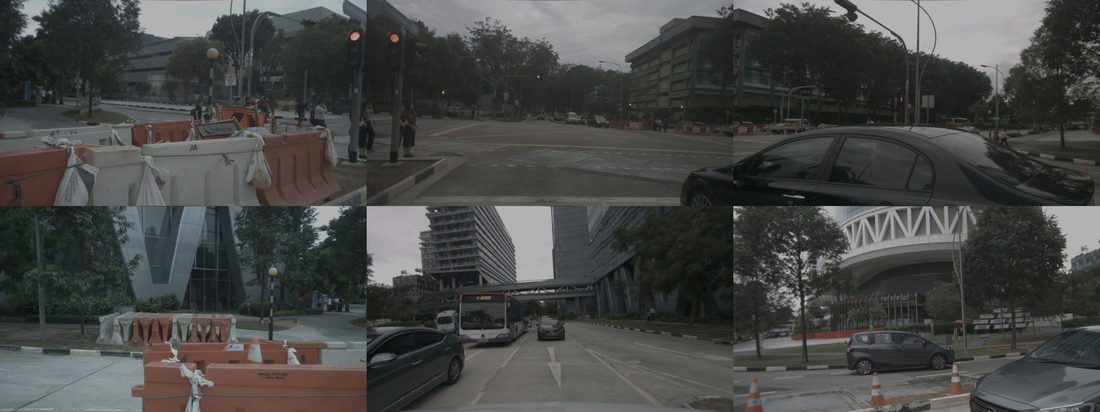} &
\includegraphics[width=\linewidth]{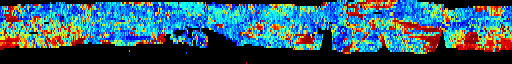} \\
\end{tabular}

\scriptsize (b) Brightness Shift \\[2pt]

\begin{tabular}{@{}m{2.8cm} m{4.7cm}@{}}
\includegraphics[width=\linewidth]{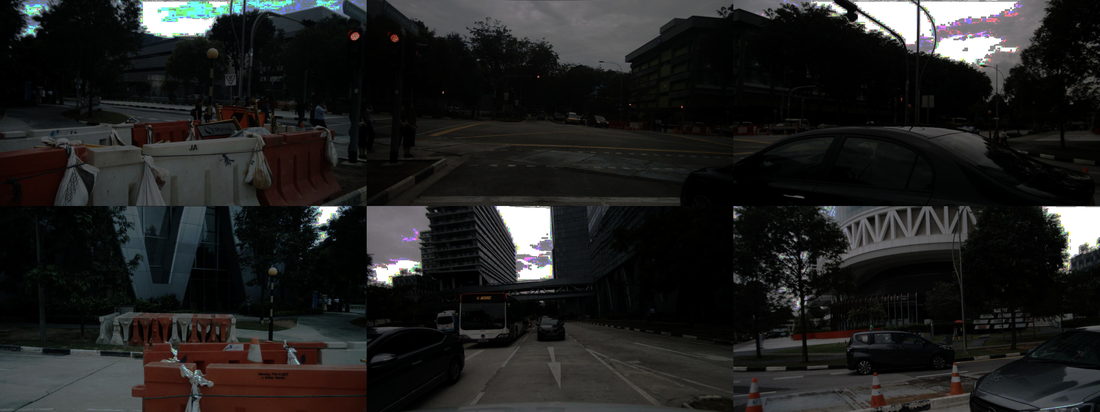} &
\includegraphics[width=\linewidth]{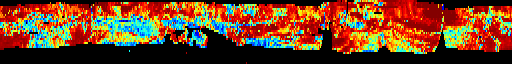} \\
\end{tabular}

\scriptsize (c) Out-of-Domain (Dark Zurich~\cite{sakaridis2019guided}) \\[2pt]

\begin{tabular}{@{}m{3.0cm} m{4.7cm}@{}}
\includegraphics[width=\linewidth]{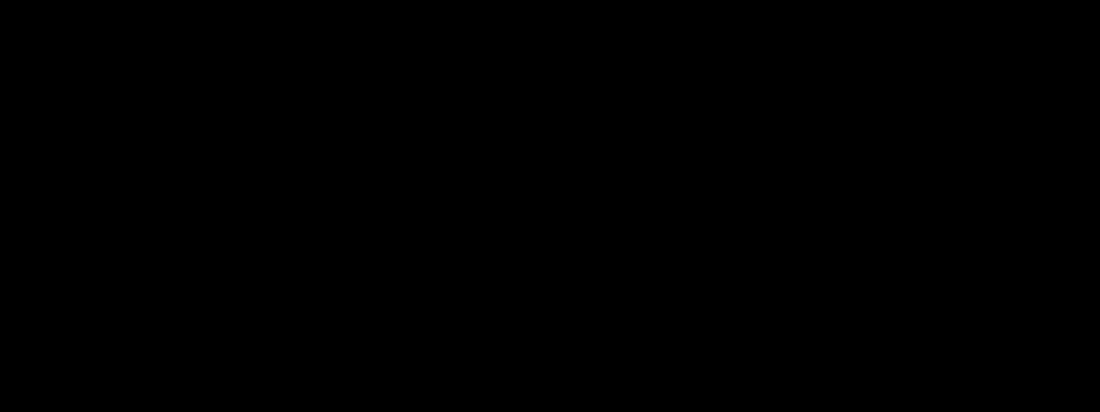} &
\includegraphics[width=\linewidth]{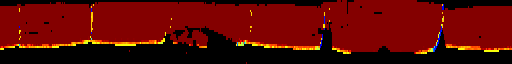} \\
\end{tabular}

\scriptsize (d) Sensor Dropout \\[2pt]

\caption{Illustration of our uncertainty module on a Panoptic nuScenes sample. The right column shows predicted uncertainty under increasing synthetic degradations. The jet colormap marks low uncertainty in \textbf{\textcolor{blue}{blue}} and high uncertainty in \textbf{\textcolor{red}{red}}. Mild distortions (b) keep uncertainty low, while strong distortions (c) and sensor dropout (d) produce high uncertainty. Black regions indicate areas without a camera to range-view (RV) mapping.}\label{fig:augmentations}
\vspace{-0.3cm}
\end{figure}

\subsection{Uncertainty-Aware Fusion Module}\label{sec:fusion}
At each scale $s \in \{4, 8, 16, 32\}$, the fusion module combines range-view (RV) aligned camera features $\mathbf{F}_{C,s}$ and LiDAR features $\mathbf{F}_{L,s}$ through uncertainty-guided cross-modal interaction to produce fused features $\mathbf{F}_{F,s}$. 

{\parskip=2pt
\noindent\textbf{Uncertainty Quantification of Camera Features}: 
We model uncertainty in camera features as instability under input degradations. Reliable features remain consistent across mild corruptions, whereas features from degraded inputs (e.g., underexposed or overexposed images) exhibit higher variability. To learn this relationship, we train a lightweight 3-layer MLP, $\mathcal{U}_{\theta, s}$, to regress feature instability. 
During training, for each image $\mathbf{I}_{\text{orig}}$, we sample a corrupted counterpart $\mathbf{I}_{\text{aug}}$ using a diverse set of non-spatial augmentations such as brightness and contrast shifts, sensor dropout, and cross-domain histogram matching. The latter adjusts image color statistics to match those from Cityscapes~\cite{cordts2016cityscapes}, COCO~\cite{lin2014microsoft}, and Dark Zurich~\cite{sakaridis2019guided}, thereby introducing out-of-domain photometric variations. With probability of 0.1, we set $\mathbf{I}_{\text{aug}} = \mathbf{I}_{\text{orig}}$ to provide stable zero-uncertainty samples. 
Both $\mathbf{I}_{\text{orig}}$ and $\mathbf{I}_{\text{aug}}$ are processed by the frozen encoder and view transformation (\cref{sec:image_encoding_transformation}) to obtain multi-scale, RV-aligned camera features $\mathbf{F}_{C,s}^{\text{orig}}$ and $\mathbf{F}_{C,s}^{\text{aug}}$.}

We define the ground-truth instability $\mathbf{d}_{\text{gt},s}$ as the L2 norm between original and augmented features at each spatial location $(u,v)$:
\begin{equation}
    \mathbf{d}_{\text{gt},s}(u,v) = \left\| \mathbf{F}_{C,s}^{\text{orig}}(u,v) - \mathbf{F}_{C,s}^{\text{aug}}(u,v) \right\|_2.
\end{equation}
The MLP $\mathcal{U}_{\theta, s}$ is trained to predict this instability from the augmented features:
\begin{equation}
    \mathbf{d}_{\text{pred},s} = \mathcal{U}_{\theta, s}(\mathbf{F}_{C,s}^{\text{aug}}).
\end{equation}
It learns to associate corrupted feature patterns with their corresponding magnitude of instability.
The network is optimized using a Huber loss, which provides robustness to outliers:
\begin{equation}
    \mathcal{L}_{\text{unc}} = \frac{1}{N} \sum_{(u,v),s} \mathcal{L}_{\delta}\!\left(\mathbf{d}_{\text{pred},s}(u,v) - \mathbf{d}_{\text{gt},s}(u,v)\right),
\end{equation}
where $N$ is the number of spatial locations and $\mathcal{L}_{\delta}$ is the Huber loss function with threshold $\delta$, set to $1.0$:
\begin{equation}
\mathcal{L}_{\delta}(a) = 
\begin{cases} 
    0.5 a^2 & \text{if } |a| \le \delta, \\
    \delta (|a| - 0.5 \delta) & \text{otherwise}.
\end{cases}
\end{equation}

Finally, the predicted instability $\mathbf{d}_{\text{pred},s}$ is transformed into a uncertainty score $\mathbf{U}_{C,s} \in [0,1]$:
\begin{equation}
    \mathbf{U}_{C,s} = 1 - \exp(-\mathbf{d}_{\text{pred},s}).
    \label{eq:uncertainty_score}
\end{equation}
This uncertainty score modulates the cross-modal interaction within the fusion module. 
Examples of augmentations and corresponding uncertainty maps for $s{=}4$ are shown in \cref{fig:augmentations} and further discussed in the supplementary material.

{\parskip=2pt
\noindent\textbf{Uncertainty-Guided Fusion}: 
We propose an uncertainty-guided fusion mechanism that adaptively integrates LiDAR and camera features through cross-modal interaction. At each scale $s$, given LiDAR features $\mathbf{F}_{L,s}$ and camera features $\mathbf{F}_{C,s}$, fusion is achieved via a deformable attention~\cite{zhu2020deformable} that aggregates spatially aligned visual information conditioned on LiDAR queries:}
\begin{equation}
    \mathbf{F}_{A,s} = \sum_{l=1}^{L} \sum_{p=1}^{P} A_{l,p} \, \tilde{\mathbf{F}}_{C,s}^{(l,p)},
\end{equation}
where $A_{l,p}$ are query-dependent attention weights measuring the relevance of each sampled camera feature $\mathbf{F}_{C,s}^{(l,p)}$ from the $l$-th level and $p$-th sampling point. We set $L{=}1$ attention level and $P{=}4$ sampling points per query. The uncertainty-modulated camera features are defined as
\begin{equation}
\tilde{\mathbf{F}}_{C,s}^{(l,p)} = (1 - \mathbf{U}_{C,s}) \odot \mathbf{F}_{C,s}^{(l,p)},
\end{equation}
where $\mathbf{U}_{C,s}$ denotes the predicted uncertainty map and $\odot$ represents element-wise multiplication. This formulation integrates deformable attention with uncertainty-driven weighting, allowing each LiDAR query to attend selectively to spatially relevant and reliable visual evidence. Regions with high uncertainty (e.g., overexposed or occluded areas) are attenuated, preventing them from dominating the fusion. The attended features are fused with the LiDAR representation as\looseness=-1
\begin{equation}
\mathbf{F}_{F,s} = \mathbf{F}_{L,s} + \mathbf{F}_{A,s}.
\end{equation}
This fusion preserves the geometric accuracy of LiDAR while enriching it with semantically reliable visual context. The resulting fused features $\mathbf{F}_{F,s}$ are subsequently passed to the 2D-3D hybrid decoder.

\subsection{Hybrid 2D-3D Panoptic Decoder}\label{sec:decoder}
The final component of our network is the hybrid 2D-3D panoptic decoder, which bridges 2D range-view features and 3D point-cloud outputs. It tackles two main challenges. First, directly predicting a 2D segmentation map and lifting it to 3D via $\mathcal{P}_{\text{RV} \rightarrow \text{3D}}$ is unreliable, since multiple 3D points project to the same 2D pixel $(u,v)$. Consequently, 2D predictions may propagate to occluded points, introducing label ambiguity. Second, the 360° LiDAR projection causes objects near the horizontal boundary (e.g., 0°/360°) to split across the 2D grid. A 2D-only decoder, unaware of this geometric continuity, would interpret them as separate instances, leading to fragmented predictions. To address these challenges, we propose a hybrid decoder that interleaves 2D feature processing with a 3D-aware unprojection mechanism, enabling explicit learning of 360° object continuity and resolving spatial ambiguity in 3D. Our approach follows the Mask2Former~\cite{cheng2022masked} paradigm, consisting of a pixel decoder and a transformer decoder.

{\parskip=2pt
\noindent\textbf{2D Pixel and Transformer Decoder}: 
The fused features $\mathbf{F}_{F,s}$ at scales $s \in \{4, 8, 16, 32\}$ are first processed by the pixel decoder, a multi-scale deformable attention-based feature pyramid~\cite{zhu2020deformable}. It outputs (1) multi-scale feature maps $\mathbf{M}_s$ at $\{8, 16, 32\}$ resolutions and (2) a high-resolution mask feature map $\mathbf{F}_{\text{mask}}$ at scale 4 used for mask prediction. A Transformer decoder with a 3-layer block structure, repeated $L$ times, then processes $N_q$ learnable object queries, attending to $\mathbf{M}_s$ with each layer focusing on a different scale in alternation. Within each 3-layer block, the intermediate layers use the standard 2D mask head~\cite{cheng2022masked} for auxiliary supervision, while the block's final layer employs our 3D-aware mask head to generate predictions, capturing spatial continuity in the point cloud domain.}

{\parskip=2pt
\noindent\textbf{3D-Aware Mask Head}: 
This head generates a per-point feature vector $\mathbf{f}_{\text{point},j}$ for each point $\mathbf{p}_j$ in the original point cloud $\mathbf{P}$. This is achieved through a learnable, geometry-aware feature aggregation process. The module takes three inputs: (1) the high-resolution feature map $\mathbf{F}_{\text{mask}}$, (2) the range channel of the 2D range image $\mathbf{L}_{\text{RV}}$ (from \cref{sec:lidar_projection_encoding}), which stores the minimum range $r_{(u,v)}$ at each pixel, and (3) the $N$ 3D points, each with its true 3D range $r_{j,\text{true}}$ and its $\mathcal{P}_{\text{RV} \rightarrow \text{3D}}$ mapping from pixel $(u_j, v_j)$. For each point $\mathbf{p}_j$, a $K \times K$ search window is defined in the range channel centered at $(u_j, v_j)$. We then compute the absolute range difference $|r_{j,\text{true}} - r_{(u',v')}|$ between the point's true 3D range and the range value of every pixel $(u',v')$ in the $K \times K$ window. The $K$ pixels in this window with the smallest range difference are selected as 3D-consistent neighbors. The corresponding $K$ features from $\mathbf{F}_{\text{mask}}$ are gathered, concatenated, and fused through a lightweight 2-layer MLP to yield a context-aware feature $\mathbf{f}_{\text{point},j} \in \mathbb{R}^D$, where $D$ is the feature channel dimension. The resulting per-point feature map $\mathbf{F}_{\text{point}} \in \mathbb{R}^{D \times N}$ is thus a spatially consistent, 3D-aware representation, constructed from aggregation over all $N$ points, that mitigates label bleeding and instance fragmentation.\looseness=-1}

{\parskip=2pt
\noindent\textbf{Panoptic Prediction and Loss}: 
The final transformer decoder layer produces the refined query embeddings $\mathbf{Q}_{\text{embed}}$ and the per-point feature map $\mathbf{F}_{\text{point}}$. Two lightweight heads generate the final outputs. The \textit{class head} applies a linear projection on $\mathbf{Q}_{\text{embed}}$ to predict class logits $\mathbf{C} \in \mathbb{R}^{N_q \times (C_{\text{th}} + C_{\text{st}} + 1)}$, where $C_{\text{th}}$ and $C_{\text{st}}$ denote the number of thing and stuff classes, respectively, and the additional class corresponds to the no-object label. The \textit{mask head} transforms $\mathbf{Q}_{\text{embed}}$ into mask embeddings $\mathbf{E}_{\text{mask}} \in \mathbb{R}^{N_q \times D}$ through an MLP, and computes the 3D mask logits via a dot product with the per-point features:}
\begin{equation}
\mathbf{M}_{\text{3D}}(q, j) = \mathbf{E}_{\text{mask}}(q, :) \cdot \mathbf{F}_{\text{point}}(:, j).
\end{equation}
The decoder thus outputs the class logits $\mathbf{C}$ and per-point mask logits $\mathbf{M}_{\text{3D}} \in \mathbb{R}^{N_q \times N}$. Following~\cite{cheng2022masked}, we employ a set-based loss computed on a subset of sampled points for efficiency. For each query, $N_p$ points are selected using a mixture of importance sampling (points with highest uncertainty) and random sampling. Bipartite matching assigns ground-truth instances to queries, and the final loss is defined as\looseness=-1
\begin{equation}
\mathcal{L}_{\text{panoptic}} = \lambda_{\text{cls}}\mathcal{L}_{\text{cls}} + \lambda_{\text{mask}}\mathcal{L}_{\text{mask}} + \lambda_{\text{dice}}\mathcal{L}_{\text{dice}},
\end{equation}
where $\mathcal{L}_{\text{cls}}$, $\mathcal{L}_{\text{mask}}$, and $\mathcal{L}_{\text{dice}}$
are the classification, mask, and Dice losses, respectively.
This formulation enables supervision over dense 3D point predictions within the hybrid 2D-3D framework. 

\section{Experimental Evaluation}
In this section, we present a comprehensive evaluation of UP-Fuse. We first introduce the new Panoptic Waymo benchmark and the evaluation metrics in \cref{sec:dataset}. We then detail our implementation in \cref{sec:implementation_details}. Finally, we present comprehensive evaluation results, including benchmarking (\cref{sec:benchmark}),
robustness analysis (\cref{sec:robustness}), ablation studies (\cref{sec:ablation}), and qualitative results (\cref{sec:qualitative}) . 

\subsection{Datasets and Evaluation Metrics}\label{sec:dataset}

We evaluate UP-Fuse on Panoptic nuScenes~\cite{fong2022panoptic}, SemanticKITTI~\cite{behley2019semantickitti}, and our newly introduced Panoptic Waymo benchmark derived from the Waymo Open Dataset~\cite{sun2020scalability}.

{\parskip=2pt
\noindent\textbf{Panoptic Waymo}: While Panoptic nuScenes offers a 360° benchmark, its 32-beam LiDAR is comparatively sparse. To provide a more challenging, high-resolution benchmark for fine-grained fusion, we introduce panoptic annotations for the Waymo Open Dataset (WOD)~\cite{sun2020scalability}. WOD includes a dense 64-beam LiDAR and five high-resolution cameras, covering more than 180° of the scene, which makes it well-suited for multi-modal evaluation. The dataset officially provides 3D semantic segmentation and 3D bounding box labels. We generate our Panoptic Waymo ground truth by following the protocol of Panoptic nuScenes~\cite{fong2022panoptic} to merge these annotations. We retain the original 798 training scenes and 202 validation scenes, resulting in a benchmark with 15 \textit{stuff} classes and 6 \textit{thing} classes. A detailed statistical overview is presented in the supplementary material. Since WOD is approximately 3$\times$ denser than nuScenes, we exclude thing instances with fewer than 50 points (15 in Panoptic nuScenes) to match the higher point cloud fidelity.}

{\parskip=2pt
\noindent\textbf{Evaluation Metrics}:
We evaluate our model using standard panoptic segmentation metrics~\cite{kirillov2019panoptic}: Panoptic Quality (PQ), Segmentation Quality (SQ), and Recognition Quality (RQ). Furthermore, we report the modified Panoptic Quality metric ($\text{PQ}^{\dagger}$) introduced in~\cite{porzi2019seamless}.
All metrics are reported as averages over all classes,
with additional breakdowns for \textit{thing} (PQ$^\text{th}$) and \textit{stuff} (PQ$^\text{st}$) categories.}

\subsection{Implementation Details}\label{sec:implementation_details}
{\parskip=2pt
\noindent\textbf{Architecture and Pre-training}: 
We use Swin-B~\cite{liu2021swin} backbones for both LiDAR and camera. The camera encoder is pre-trained using a camera-only variant of our model, where LiDAR is used solely for view transformation. Its weights remain frozen during all multi-modal experiments. Our hybrid 2D-3D panoptic decoder operates with $N_q = 300$ queries and $L = 2$ blocks. The 3D-aware mask head aggregates features from $K = 5$ neighbors (see \cref{sec:decoder}). Further details are presented in the supplementary material.}

{\parskip=2pt
\noindent\textbf{Training Details}: 
All models are optimized using AdamW~\cite{loshchilov2017decoupled} with a base learning rate of $10^{-4}$. We apply standard spatial augmentations, including rotation, scaling, and horizontal flipping, to both modalities. For uncertainty training, we use diverse non-spatial camera corruptions such as photometric changes, sensor dropout, and out-of-domain histogram matching. The full list is detailed in the supplementary material. On Panoptic nuScenes, we use a range-view input of $32 \times 1024$ (resized to $256 \times 2048$), a batch size of $16$, and train for $80$ epochs with a $10\times$ learning rate decay applied at epochs $72$ and $76$. On Panoptic Waymo, we use a range-view input of $64 \times 2560$ (resized to $256 \times 4096$), a batch size of $8$, and train for $48$ epochs with learning rate decays at epochs $40$ and $44$. For SemanticKITTI, we use a range-view input of $64 \times 2048$ (resized to $256 \times 4096$), a batch size of $4$, and train for $24$ epochs with $10\times$ learning rate decays at epochs $16$ and $20$.\@
The loss weights $(\lambda_{\text{cls}}, \lambda_{\text{dice}}, \lambda_{\text{mask}}, \lambda_{\text{unc}})$ are set to $(5, 5, 100, 1)$ for Panoptic nuScenes and $(2, 5, 50, 1)$ for both Panoptic Waymo and SemanticKITTI.\@ During training, we sample $N_p = 12544$ points for Panoptic nuScenes and $N_p = 25088$ points for Panoptic Waymo and SemanticKITTI.\@ Camera images are resized to $256 \times 704$ pixels for Panoptic nuScenes similar to~\cite{mohan2026forecastocc} and Panoptic Waymo, and to $360 \times 640$ pixels for SemanticKITTI.\@ Lastly, FPS is measured on a single NVIDIA A40 GPU with batch size 1 following the evaluation protocol used by IAL~\cite{pan2025images}.
}

\begin{table}
    \setlength{\tabcolsep}{0.0085\linewidth}
    \centering
    \footnotesize
    \caption{Comparison of 3D panoptic segmentation performance on the Panoptic nuScenes validation set.}\label{tab:mainresults_nuval}

    \begin{tabular}{l|c|cccc|cc|c}
        \toprule
        Method & Modality & PQ & PQ$^\dagger$ & SQ & RQ & PQ$^{\text{th}}$ & PQ$^{\text{st}}$ & FPS \\
        \midrule
        CPSeg HR~\cite{li2021cpseg} & L & 71.1 & 75.6 & 82.5 & 85.5 & 71.5 & 70.6 & --- \\
        Panoptic-FusionNet~\cite{song2024panoptic} & L & 72.7 & 75.4 & 86.4 & 84.8 & 71.2 & 75.1 & --- \\
        LCPS~\cite{zhang2023lidar} & L & 72.9 & 77.6 & 88.4 & 82.0 & 72.8 & 73.0 & --- \\
        Panoptic-PHNet~\cite{li2022panoptic} & L & 74.7 & 77.7 & 88.2 & 84.2 & 74.0 & 75.9 & --- \\
        PUPS~\cite{su2023pups} & L & 74.7 & 77.3 & 89.4 & 83.3 & 75.4 & 73.6 & --- \\
        \textbf{UP-Fuse (Ours)} & L & 74.9 & 78.2 & 87.8 & 85.1 & 75.1 & 74.5 & --- \\
        CFNet~\cite{li2023center} & L & 75.1 & 78.0 & 88.8 & 84.6 & 74.8 & 76.6 & --- \\
        P3Former~\cite{xiao2025position} & L & 75.9 & 78.9 & 89.7 & 84.7 & 76.9 & 75.4 & --- \\
        CenterLPS~\cite{mei2023centerlps} & L & 76.4 & 79.2 & 86.2 & 88.0 & 77.5 & 74.6 & --- \\
        IAL~\cite{pan2025images} & L & 77.0 & 79.6 & 90.2 & 85.1 & 77.8 & 75.7 & --- \\
        \midrule
        Panoptic-FusionNet~\cite{song2024panoptic} & LC & 77.2 & 79.3 & 87.8 & 87.2 & 77.5 & 76.2 & 2.5 \\
        LCPS~\cite{zhang2023lidar} & LC & 79.8 & 84.0 & 89.8 & 88.5 & 82.3 & 75.6 & 1.7 \\
        IAL~\cite{pan2025images} & LC & 80.3 & 82.8 & 91.0 & 87.9 & --- & --- & 0.9 \\
        \textbf{UP-Fuse (Ours)} & LC & 80.7 & 84.3 & 90.3 & 89.0 & 83.0 & 77.0 & \textbf{5.7} \\
        IAL-PieAug~\cite{pan2025images} & LC & \textbf{82.3} & \textbf{84.7} & \textbf{91.5} & \textbf{89.7} & \textbf{85.3} & \textbf{77.3} & 0.9 \\
        \bottomrule
    \end{tabular}
    \vspace{-0.5em}
\end{table}

\subsection{Benchmarking Results}\label{sec:benchmark}
We evaluate our UP-Fuse approach against published state-of-the-art methods. We first report results on the primary Panoptic nuScenes benchmark, followed by SemanticKITTI, and finally on our new Panoptic Waymo benchmark.

{\parskip=2pt
\noindent\textbf{Results on Panoptic nuScenes}: As shown in \cref{tab:mainresults_nuval}, UP-Fuse (LC) reaches 80.7\% PQ on the validation split,
outperforming multi-modal baselines such as LCPS~\cite{zhang2023lidar} by 0.9\% in PQ and Panoptic-FusionNet~\cite{song2024panoptic} by 3.5\% in PQ.\@ It further yields a gain of 5.8\% in PQ over our strong LiDAR-only (L) variant (74.9\%),
which removes the image encoder and uncertainty-driven fusion module. IAL~\cite{pan2025images}, when trained with its proposed modality-synchronized augmentation (IAL-PieAug),
achieves the highest absolute performance at 82.3\% PQ.\@
Without PieAug, the base IAL fusion architecture attains 80.3\% PQ,
0.4\% lower in PQ than UP-Fuse under the same augmentation protocol.
Crucially, the competitive performance of UP-Fuse is achieved with substantially higher efficiency:
it operates at 5.7 FPS, approximately $6\times$ faster than IAL (0.9 FPS),
$3\times$ faster than LCPS, and over $2\times$ faster than Panoptic-FusionNet. This demonstrates that the previously unexplored unified 2D range-view space for multi-modal 3D panoptic segmentation
enables an effective trade-off between accuracy and runtime efficiency. Beyond standard benchmarks, we further show in \cref{sec:robustness} that our fusion framework exhibits improved robustness under sensor dropout, calibration drift, and visual domain shift.

The metric breakdown shows that the major contribution comes from \textit{thing} classes, with UP-Fuse achieving a 7.9\% boost in PQ$^{\text{th}}$ compared to the LiDAR-only variant, with an additional 2.5\% gain in $\mathrm{PQ}^{\text{st}}$ for \textit{stuff} classes. This suggests that our uncertainty-guided fusion reliably exploits visual texture cues to disambiguate sparse and confusing \textit{thing} instances in LiDAR data, while also better delineating \textit{stuff} regions with similar geometry. A similar trend is observed on the test set (\cref{tab:mainresults_nutest}), where our model achieves 81.1\% in the PQ score.}

{\parskip=2pt
\noindent\textbf{Results on SemanticKITTI}: We further validate our approach on the SemanticKITTI validation set (\cref{tab:mainresults_skittival}), where the frontal camera setup limits visual overlap with LiDAR data. Under this constrained sensing setup, IAL-PieAug~\cite{pan2025images} achieves the highest PQ at 63.1\%,
while UP-Fuse (LC) attains 61.8\% PQ, outperforming LCPS~\cite{zhang2023lidar} by 2.8\% in PQ.\@ This indicates that UP-Fuse can effectively exploit complementary visual cues when available, even under limited camera coverage.}

{\parskip=2pt
\noindent\textbf{Results on Panoptic Waymo} bridges the gap between Panoptic nuScenes,
with full surround cameras and sparse LiDAR,
and SemanticKITTI, with frontal cameras and denser LiDAR.\@
As shown in \cref{tab:mainresults_waymoval}, UP-Fuse (LC) achieves 60.9\% PQ,
a gain of 3.5\% in PQ over its LiDAR-only baseline.
This performance also surpasses other multi-modal methods,
outperforming IAL~\cite{pan2025images} by 0.5\% in PQ.\@
The training protocols for all baselines are provided in the supplementary material.
}

\begin{table}
    \setlength{\tabcolsep}{0.010\linewidth}
    \centering
    \footnotesize
    \caption{Comparison of 3D panoptic segmentation results on the Panoptic nuScenes test set.}\label{tab:mainresults_nutest}

    \begin{tabular}{l|c|cccc|cc}
        \toprule
        Method & Modality & PQ & PQ$^\dagger$ & SQ & RQ & PQ$^\text{th}$ & PQ$^\text{st}$   \\
        \midrule
         MaskPLS~\cite{marcuzzi2023mask} & L & 61.1 & 64.3 & 86.8 & 68.5 & 54.3  & 72.4  \\
        EfficientLPS~\cite{sirohi2021efficientlps} & L & 62.4 & 66.0 & 83.7 & 74.1 & 57.2 & 71.1 \\
        Panopitc-PolarNet~\cite{zhou2021panoptic} & L & 63.6 & 67.1 & 84.3 & 75.1 & 59.0 & 71.3 \\
        LCPS~\cite{zhang2023lidar} & L & 72.8 & 76.3 & 88.6 & 81.7 & 72.4 & 73.5\\
        CPSeg~\cite{li2021cpseg} & L & 73.2 & 76.3 & 88.1 & 82.7 & 72.9 & 74.0\\
        Panoptic-PHNet~\cite{zhou2021panoptic} & L & 80.1 & 82.8 & 91.1 & 87.6 & 82.1 & 76.6\\
        \midrule
        LCPS~\cite{zhang2023lidar} & LC & 79.5 & 82.3 & 90.3 & 87.7 & 81.7 & 75.9\\
        \textbf{UP-Fuse (Ours)} & LC & 81.1 & 83.4 & 91.3 & 88.5 & 83.6 & 76.9\\
        IAL-PieAug~\cite{pan2025images} & LC & \textbf{82.0} & \textbf{84.3} & \textbf{91.6} & \textbf{89.3} & \textbf{84.8} & \textbf{77.5}
        \\ 
        \bottomrule
    \end{tabular}
\end{table}

\begin{table}
    \setlength{\tabcolsep}{0.010\linewidth}
    \centering
    \footnotesize
    \caption{Comparison of 3D panoptic segmentation results on the SemanticKITTI validation set.}\label{tab:mainresults_skittival}

    \begin{tabular}{l|c|cccc}
        \toprule
        Method & Modality & PQ & PQ$^\dagger$ & SQ & RQ  \\
        \midrule
        LCPS~\cite{zhang2023lidar} & L & 55.7 & 65.2 & 65.8 & 74.0 \\
        EfficientLPS~\cite{sirohi2021efficientlps} & L & 59.2 & 65.1 & 69.8 & 75.0 \\
        \textbf{UP-Fuse (Ours)} & L & 60.4 & 67.8 & 74.9 & 71.7 \\
        Panoptic-PHNet~\cite{li2022panoptic} & L & 61.7 & --- & --- & --- \\
        IAL~\cite{pan2025images} & L & 62.0 & 65.1 & 76.0 & 71.9 \\
        CenterLPS~\cite{mei2023centerlps} & L & 62.1 & 67.0 & 72.0 & \textbf{80.7} \\
        P3Former~\cite{xiao2025position} & L & 62.6 & 66.2 & 72.4 & 76.2 \\
        GP-S3Net~\cite{razani2021gp} & L & 63.3 & \textbf{71.5} & 81.4 & 75.9 \\
        PUPS~\cite{su2023pups} & L & \textbf{64.4} & 68.6 & \textbf{81.5} & 74.1 \\
        \midrule
        LCPS~\cite{zhang2023lidar} & LC & 59.0 & 68.8 & 68.9 & 79.8 \\
        \textbf{UP-Fuse (Ours)} & LC & 61.8 & 69.1 & 75.2 & 73.1 \\
        IAL-PieAug~\cite{pan2025images} & LC & 63.1 & 66.3 & 81.4 & 72.9 \\
        \bottomrule
    \end{tabular}

\end{table}

\begin{table}
    \setlength{\tabcolsep}{0.010\linewidth}
    \centering
    \footnotesize
    \caption{Comparison of 3D panoptic segmentation results on the Panoptic Waymo validation set.}\label{tab:mainresults_waymoval}

    \begin{tabular}{l|c|cccc|cc}
        \toprule
        Method & Modality & PQ & PQ$^\dagger$ & SQ & RQ & PQ$^\text{th}$ & PQ$^\text{st}$ \\
        \midrule
        Panoptic-FusionNet~\cite{song2024panoptic} & L & 53.9 & 63.2 & 77.4 & 68.2 & 55.1 & 53.4 \\
        LCPS~\cite{zhang2023lidar} & L & 54.9 & 63.1 & 77.8 & 69.2 & 58.2 & 53.6 \\
        EfficientLPS~\cite{sirohi2021efficientlps} & L & 56.3 & 64.5 & 78.3 & 70.6 & 59.7 & 54.9 \\
        \textbf{UP-Fuse (Ours)} & L & 57.4 & 65.1 & 78.8 & 71.5 & 61.3 & 55.8 \\
        IAL~\cite{pan2025images} & L & 58.2 & 66.1 & 79.2 & 72.1 & 62.5 & 56.4 \\
        P3Former~\cite{xiao2025position} & L & 59.3 & 67.4 & 79.3 & 73.5 & 63.8 & 57.5 \\
        \midrule
        Panoptic-FusionNet~\cite{song2024panoptic} & LC & 56.2 & 64.8 & 78.5 & 70.9 & 58.7 & 55.2 \\
        LCPS~\cite{zhang2023lidar} & LC & 58.9 & 66.7 & 79.3 & 72.8 & 63.1 & 57.2 \\
        IAL~\cite{pan2025images} & LC & 60.4 & 68.2 & 79.4 & 74.8 & 65.3 & 58.4 \\
        \textbf{UP-Fuse (Ours)} & LC & \textbf{60.9} & \textbf{69.0} & \textbf{79.6} & \textbf{75.4} & \textbf{66.7} & \textbf{58.6} \\
        \bottomrule
    \end{tabular}
\end{table}

\subsection{Robustness Analysis}\label{sec:robustness}
Real-world robotic deployment requires resilience to three failure modes~\cite{koopman2016challenges}:
complete sensor loss (dropout), mechanical misalignment (calibration drift), and environmental degradation (visual domain shift).
We evaluate UP-Fuse under these conditions to assess its reliability
for safety-critical robotic perception. Additionally, \cref{fig:qual_robustness} qualitatively compares IAL and UP-Fuse under these failure modes.

{\parskip=2pt
\noindent\textbf{Robustness to Sensor Dropout}: 
We evaluate robustness under complete camera failure, with results reported in \cref{tab:nuscenes_robustness}. To ensure a fair comparison, all multi-modal baselines are retrained from scratch using the same camera dropout augmentations applied in UP-Fuse. In the full modality setting (LC), all fusion-based methods achieve a substantial PQ improvement over their LiDAR-only counterpart (L), with LCPS~\cite{zhang2023lidar} obtaining the largest gain of 6.4\% in PQ, followed by 5.8\% from our UP-Fuse. The critical evaluation setting for robustness is when the camera input is removed at inference (L*). In this regime, the limitations of fusion strategies that focus solely on relevance rather than reliability become evident. Panoptic-FusionNet~\cite{song2024panoptic}, LCPS~\cite{zhang2023lidar}, and IAL~\cite{pan2025images} exhibit substantial degradation, with reductions of 5.0\%, 4.2\%, and 4.6\% in PQ, respectively, all falling below their own L-only baselines. UP-Fuse demonstrates a markedly different response. Guided by its uncertainty-driven fusion module, the model suppresses corrupted image features and relies more heavily on the LiDAR representation. Its performance decreases by only 1.2\% in PQ, remaining close to its strong L-only baseline. These findings indicate that robustness cannot be obtained through data augmentation alone and instead requires an architecture capable of adaptively modulating unreliable inputs.}

\begin{figure}
  \centering
  \includegraphics[width=\linewidth]{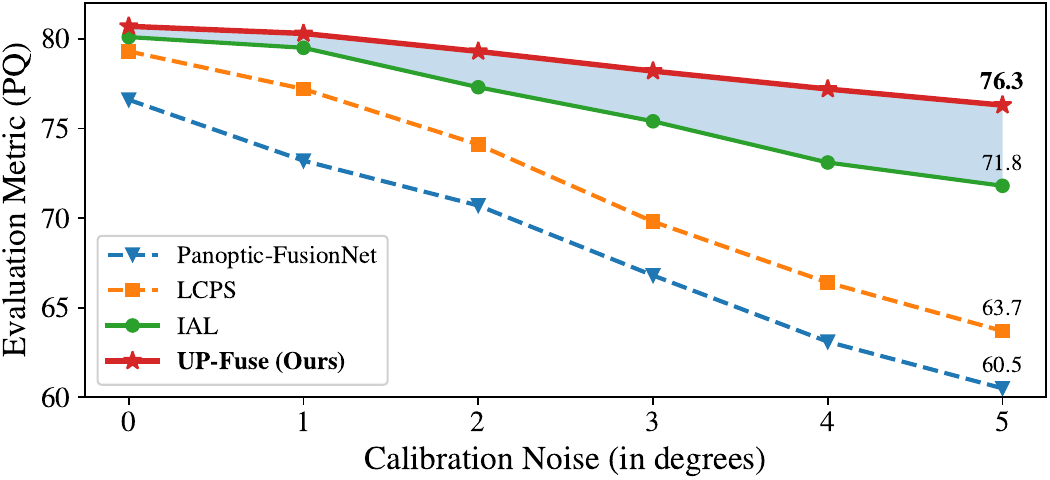}
    \caption{Robust performance comparison on the Panoptic nuScenes validation set under increasing calibration drift between LiDAR and camera over rotation magnitudes from $0^{\circ}$ to $5^{\circ}$. UP-Fuse (red) outperforms all baselines, dropping only $4.4\%$ in PQ compared to $>8\%$ for state-of-the-art methods. Refer to the supplementary material for visualization of the projection shifts.}\label{fig:plot_calib}
\end{figure}

{\parskip=2pt
\noindent\textbf{Robustness to Calibration Drift}:
We evaluate geometric robustness by simulating mechanical calibration drift through random rotational perturbations $\theta \in [0^{\circ}, 5^{\circ}]$ applied to the LiDAR-camera extrinsics at inference. As shown in \cref{fig:plot_calib}, all multi-modal baselines exhibit increasing performance degradation as misalignment increases. While IAL~\cite{pan2025images} improves over Panoptic-FusionNet~\cite{song2024panoptic} and LCPS~\cite{zhang2023lidar}, it still incurs an 8.3\% drop in PQ at $5^{\circ}$. In contrast, UP-Fuse limits the degradation to just 4.4\%. These results confirm the adaptive behavior of UP-Fuse: as cross-modal alignment deteriorates, the uncertainty-driven fusion module progressively reduces the influence of unreliable image features.}

{\parskip=2pt
\noindent\textbf{Robustness to Visual Domain Shift}: 
We further assess robustness under a visual domain shift from daytime to nighttime conditions. To this end, all models are retrained using the 630 daytime scenes from Panoptic nuScenes and evaluated on 15 held-out nighttime scenes. This setting represents a naturally occurring domain shift, where camera observations are substantially degraded while LiDAR data remain reliable.
As shown in \cref{tab:nuscenes_day2night}, all baseline fusion methods are highly sensitive to this shift.
Having learned to rely on visual cues under daytime illumination, Panoptic-FusionNet~\cite{song2024panoptic},
LCPS~\cite{zhang2023lidar}, and IAL~\cite{pan2025images} incorrectly fuse dark and uninformative image features,
resulting in drops of 3.1\%, 2.7\% and 2.1\% in PQ relative to their LiDAR-only baselines (L), respectively.
In contrast, UP-Fuse demonstrates strong architectural robustness. Its uncertainty-driven fusion module identifies regions where visual features are unreliable and selectively incorporates only the complementary cues that remain informative. As a result, UP-Fuse maintains its performance and achieves a 0.1\% improvement in PQ, effectively reverting to its robust L-only baseline and avoiding any significant degradation. We note that the night scene evaluation lacks four \textit{thing} classes (bus, construction vehicle, trailer, and traffic cone).}

\begin{table}
    \setlength{\tabcolsep}{0.009\linewidth}
    \centering
    \footnotesize
    \caption{Robust performance comparison on the Panoptic nuScenes validation set under missing camera modality conditions. $\Delta$PQ denotes the change compared to the corresponding LiDAR-only variant.}\label{tab:nuscenes_robustness}

    \begin{tabular}{l|c|ccc|c}
        \toprule
        Method & Modality & PQ & PQ$^\text{th}$ & PQ$^\text{st}$ & $\Delta$PQ \\
        \midrule
        Panoptic-FusionNet~\cite{song2024panoptic} & L & 72.7 & 71.2 & 75.1 & --- \\
        LCPS~\cite{zhang2023lidar} & L & 72.9 & 72.8 & 73.0 & --- \\
        \textbf{UP-Fuse (Ours)} & L & 74.9 & 75.1 & 74.5 & --- \\
        IAL~\cite{pan2025images} & L & 77.0 & 77.8 & 75.7 & --- \\
        \midrule
        Panoptic-FusionNet~\cite{song2024panoptic} & LC & 76.6 & 77.1 & 76.0 & +3.9 \\
        LCPS~\cite{zhang2023lidar} & LC & 79.3 & 81.7 & 75.3 & \textbf{+6.4} \\
        IAL~\cite{pan2025images} & LC & 80.1 & 82.2 & 76.5 & +3.7 \\
        \textbf{UP-Fuse (Ours)} & LC & 80.7 & 83.0 & 77.0 & +5.8 \\
        \midrule
        Panoptic-FusionNet~\cite{song2024panoptic} & L$^*$ & 67.7 & 65.3 & 71.7 & -5.0 \\
        LCPS~\cite{zhang2023lidar} & L$^*$ & 68.7 & 67.9 & 70.1 & -4.2 \\
        IAL~\cite{pan2025images} & L$^*$ & 72.4 & 71.6 & 73.8 & -4.6 \\
        \textbf{UP-Fuse (Ours)} & L$^*$ & 73.7 & 73.6 & 74.1 & \textbf{-1.2} \\
        \bottomrule
    \end{tabular}
\end{table}

\begin{table}
    \setlength{\tabcolsep}{0.012\linewidth}
    \centering
    \footnotesize
    \caption{Comparison of robustness to visual domain shift on the Panoptic nuScenes validation set. Models trained on day scenes are evaluated on the night split to assess performance under corrupted visual features.}\label{tab:nuscenes_day2night}

    \begin{tabular}{l|c|ccc|c}
        \toprule
        Method & Modality & PQ & PQ$^\text{th}$ & PQ$^\text{st}$ & $\Delta$PQ \\
        \midrule
        Panoptic-FusionNet~\cite{song2024panoptic} & L & 63.1 & 61.3 & 64.8 & --- \\
        LCPS~\cite{zhang2023lidar} & L & 63.3 & 63.1 & 63.4 & --- \\
        \textbf{UP-Fuse (Ours)} & L & 64.2 & 63.9 & 64.4 & --- \\
        IAL~\cite{pan2025images} & L & 65.6 & 65.5 & 65.7 & --- \\
        \midrule
        Panoptic-FusionNet~\cite{song2024panoptic} & LC & 60.0 & 56.2 & 63.7 & -3.1 \\
        LCPS~\cite{zhang2023lidar} & LC & 60.6 & 58.3 & 62.9 & -2.7 \\
        IAL~\cite{pan2025images} & LC & 63.5 & 63.1 & 64.2 & -2.1 \\
        \textbf{UP-Fuse (Ours)} & LC & 64.3 & 64.9 & 63.7 & \textbf{+0.1} \\
        \bottomrule
    \end{tabular}
\end{table}

\subsection{Ablation Study}\label{sec:ablation}

{\parskip=2pt
\noindent\textbf{Architectural Ablation}: As shown in \cref{tab:arch_ablation}, we perform a bottom-up analysis of our architecture. Our first baseline (M1) adapts a 2D decoder~\cite{cheng2022masked} to the range-view and uses a KNN-based post-processing~\cite{milioto2019rangenet++} to lift 2D predictions to 3D. Replacing this with our Hybrid 2D-3D Panoptic Decoder (M2) provides a significant 2.1\% gain in PQ (74.9\% vs. 72.8\% PQ), confirming it successfully alleviates projection ambiguities in the 360° range-view. We then introduce the camera modality with concatenation fusion. We find that using dense depth (VT-$\mathcal{D}_{\text{dense}}$) provides a 0.8\% gain in PQ over sparse depth (VT-$\mathcal{D}_{\text{sparse}}$) in view transformer. Next, we replace simple concatenation with Deformable Cross-Modal Interaction (D-CMI), which provides a substantial 2.4\% boost in PQ, demonstrating the clear superiority of attention-based fusion. Adding our sensor degradation augmentations (Augs.) results in a slight drop to 78.9\% PQ, as the model learns a robust but overly-cautious representation. The drop is mitigated by enabling our Uncertainty-Aware D-CMI, which elevates performance to 80.7\% PQ. This highlights that uncertainty-awareness is essential for effectively leveraging the augmentations.}

\begin{figure}[t]
    \centering
    
    \begin{subfigure}{\linewidth}
        \centering
        \includegraphics[width=0.9\linewidth]{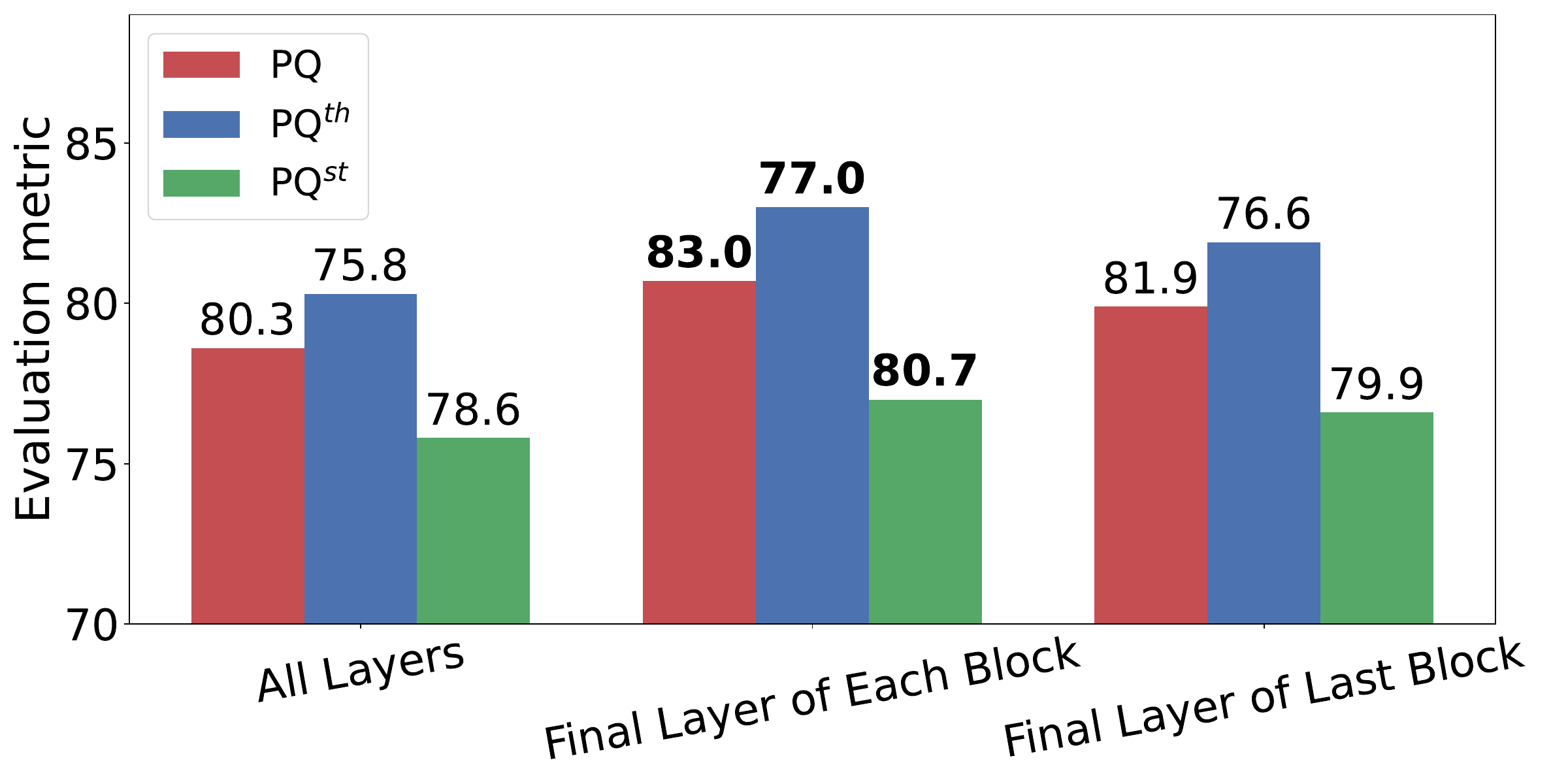}
        \caption{Analysis of Head Configuration}\label{fig:subfig_a}
    \end{subfigure}
    \hfill
    \begin{subfigure}{\linewidth}
        \centering
        \includegraphics[width=0.9\linewidth]{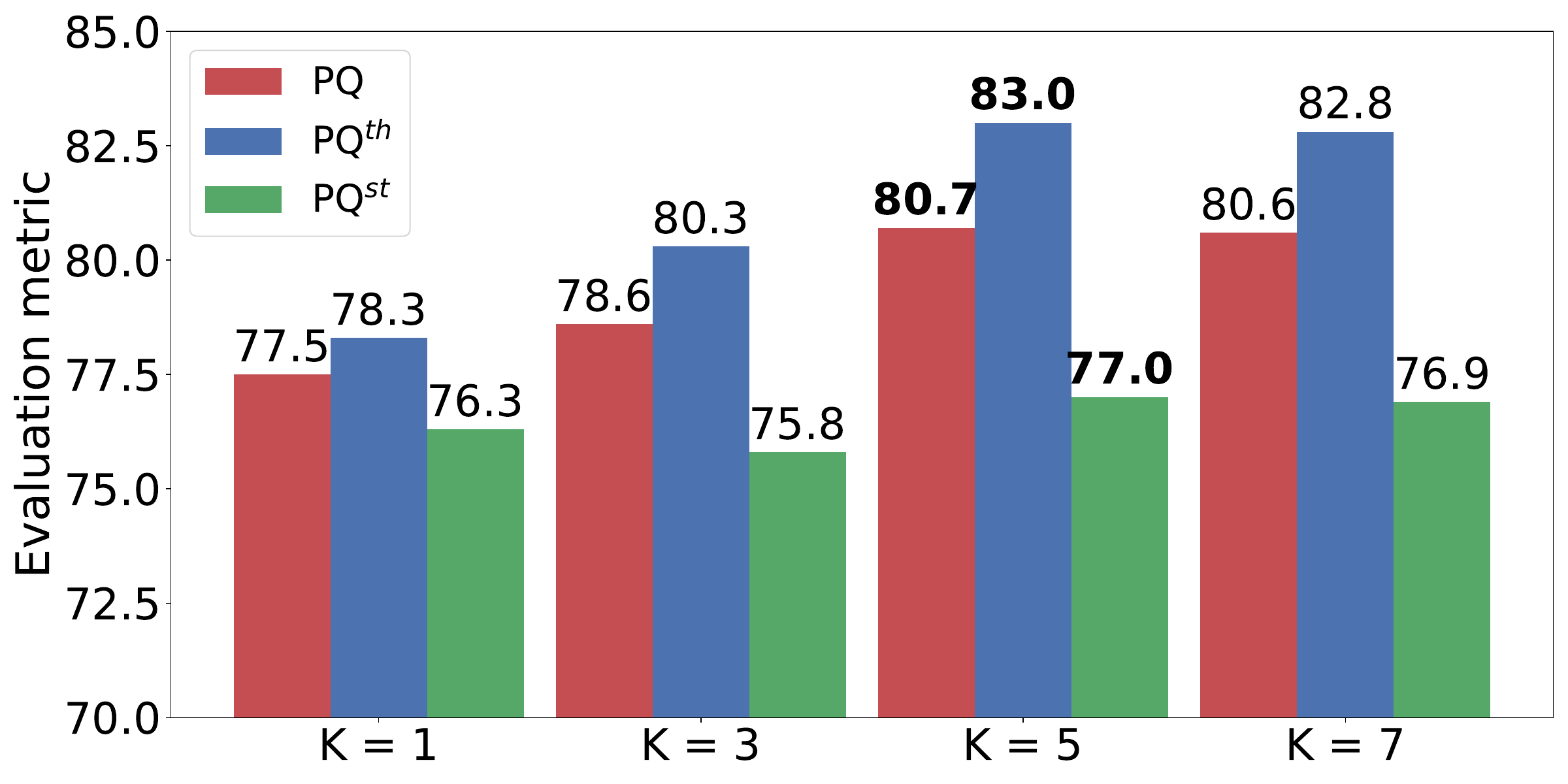}
        \caption{Impact of Neighborhood Size K}\label{fig:subfig_b}
    \end{subfigure}

    \caption{Ablation studies on the key hyperparameters of our Hybrid 2D-3D Panoptic Decoder.}\label{fig:decoder_ablation}
\end{figure}

{\parskip=2pt
\noindent\textbf{Hybrid Decoder Design}: 
In \cref{fig:decoder_ablation}, we ablate the two key hyperparameters of our Hybrid 2D-3D Panoptic Decoder. First, in \cref{fig:subfig_a}, we analyze the placement of the 3D-aware mask head within the decoder layers. We find that performing the 3D-lifting operation in the final layer of each decoder block provides the best performance (80.7\% PQ), as it allows the queries to first refine themselves in the 2D range-view before being lifted to 3D. Second, in \cref{fig:subfig_b}, we ablate the neighborhood size $K$ for the 3D-aware feature aggregation, showing that $K=5$ is the optimal choice for balancing context aggregation and feature specificity.}

{\parskip=2pt
\noindent\textbf{Uncertainty Quantification}: 
Lastly, we evaluate whether our uncertainty estimator captures visual feature instability under camera degradation. As shown in \cref{tab:pearson_correlation}, the predicted instability is consistently positively correlated with the ground-truth feature deviation across different corruptions. The correlation is highest under sensor dropout ($\rho=0.9595$), where the degradation causes severe and localized feature collapse. Strong correlations under out-of-domain shifts ($\rho=0.7802$) and fog ($\rho=0.7258$) further indicate that the estimator generalizes beyond simple missing-sensor cases. These results support our design choice of using feature-level uncertainty to identify unreliable visual cues, enabling the Uncertainty-Aware D-CMI module to down-weight degraded camera features during fusion.}

\begin{table}[t]
\centering
\footnotesize
\setlength{\tabcolsep}{2pt}
\caption{Ablation study of our UP-Fuse architecture on the Panoptic nuScenes validation set. VT denotes View-Transformation. D-CMI denotes Deformable Cross-Modal Interaction. Augs.\ denotes Augmentations.}\label{tab:arch_ablation}

\begin{tabular}{l|c|ccc}
\toprule
Model Variant & Modality & PQ & PQ$^\text{th}$ & PQ$^\text{st}$ \\
\midrule

M1 (2D-3D Post Processing~\cite{milioto2019rangenet++}) & L & 72.8 & 72.2 & 73.8 \\
M2 (Hybrid 2D-3D Decoder) & L & 74.9 & 75.1 & 74.5 \\

\midrule

M2 + Concat Fusion (VT-$\mathcal{D}_{\text{sparse}}$) & LC & 76.4 & 77.2 & 75.1 \\
M2 + Concat Fusion (VT-$\mathcal{D}_{\text{dense}}$) & LC & 77.2 & 78.4 & 75.2 \\
M2  + D-CMI (VT-$\mathcal{D}_{\text{dense}}$) & LC & 79.6 & 81.6 & 76.3 \\
\quad \quad + Sensor Degradation Augs. & LC & 78.9 & 80.7 & 75.9 \\
\quad \quad + Uncertainty-Aware D-CMI & LC & \textbf{80.7} & \textbf{83.0} & \textbf{77.0} \\

\bottomrule
\end{tabular}
\end{table}

\begin{table}[t]
\centering
\renewcommand{\arraystretch}{1.1}
\caption{Pearson correlation $\rho(d_{\text{gt}}, d_{\text{pred}})$ under camera sensor degradation on the Panoptic nuScenes validation set. The predicted instability $d_{\text{pred}}$ is consistently positively correlated with the ground-truth feature instability $d_{\text{gt}}$, with the strongest correlation under sensor dropout.}
\label{tab:pearson_correlation}
\begin{tabular}{@{}lc@{}}
    \toprule
    Degradation Type & $\rho(d_{\text{gt}}, d_{\text{pred}})$ \\
    \midrule
    Brightness Shift & 0.6907 \\
    Color Jitter     & 0.6369 \\
    Out-of-Domain    & 0.7802 \\
    Fog              & 0.7258 \\
    Sensor Dropout   & 0.9595 \\
    \bottomrule
\end{tabular}
\end{table}

\begin{figure}[t]
    \centering
    \setlength{\tabcolsep}{1pt}
    \renewcommand{\arraystretch}{0.1}
    \footnotesize

    \begin{tabular}{@{}m{0.3cm}cccc@{}}
        &
        \scriptsize Visual Context &
        \scriptsize Ground Truth &
        \scriptsize IAL (Baseline) &
        \scriptsize UP-Fuse (Ours) \\[3pt]

        \rotatebox{90}{\scriptsize (a) Sensor Dropout} &
        \begin{tabular}{@{}c@{}}
            \begin{tikzpicture}
                \node[inner sep=0] (img) {\includegraphics[width=0.22\columnwidth, height=0.11\columnwidth, keepaspectratio=false]{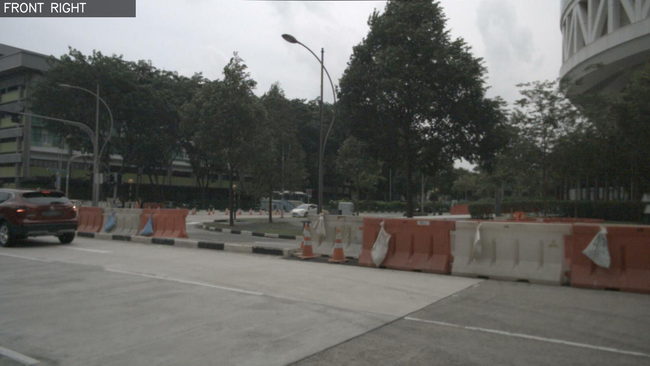}};
                \begin{scope}
                    \clip (img.south west) rectangle (img.north east);
                    \draw[red, line width=1.5pt, opacity=0.4] (img.north west) -- (img.south east);
                    \draw[red, line width=1.5pt, opacity=0.4] (img.north east) -- (img.south west);
                \end{scope}
            \end{tikzpicture}
            \tabularnewline[1pt]
            \begin{tikzpicture}
                \node[inner sep=0] (img) {\includegraphics[width=0.22\columnwidth, height=0.11\columnwidth, keepaspectratio=false]{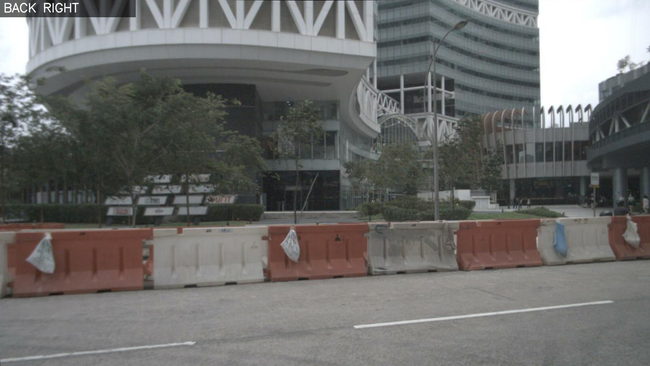}};
                \begin{scope}
                    \clip (img.south west) rectangle (img.north east);
                    \draw[red, line width=1.5pt, opacity=0.4] (img.north west) -- (img.south east);
                    \draw[red, line width=1.5pt, opacity=0.4] (img.north east) -- (img.south west);
                \end{scope}
            \end{tikzpicture}
        \end{tabular} &
        \begin{tabular}{@{}c@{}}
            \fbox{\includegraphics[width=0.21\columnwidth]{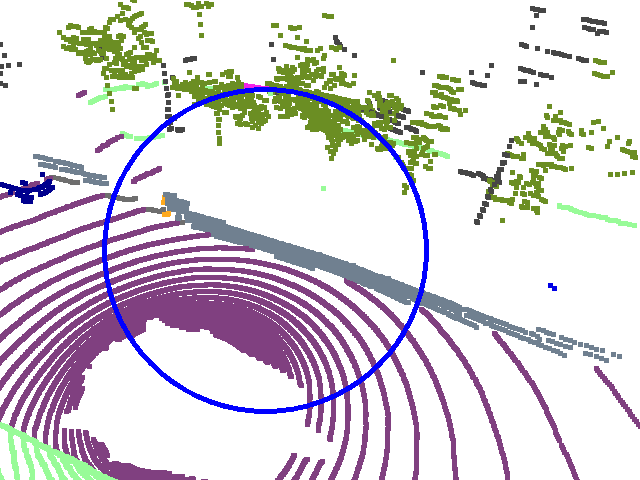}} \\[-2pt]
            \fbox{\includegraphics[width=0.21\columnwidth]{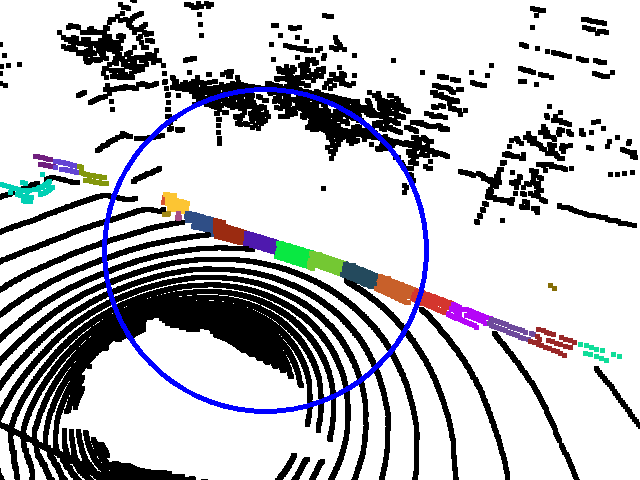}}
        \end{tabular} &
        \begin{tabular}{@{}c@{}}
            \fbox{\includegraphics[width=0.21\columnwidth]{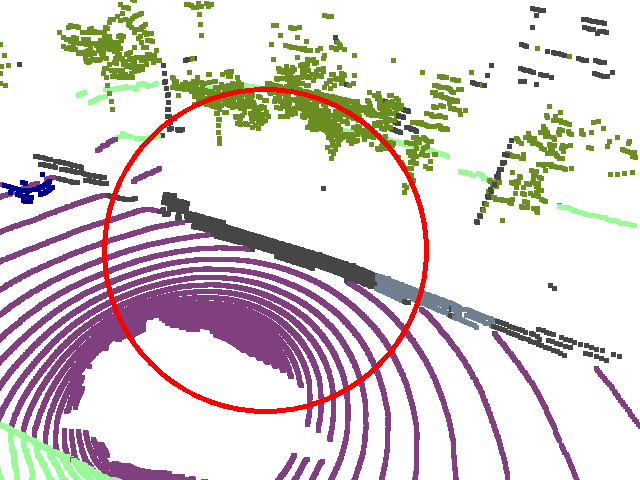}} \\[-2pt]
            \fbox{\includegraphics[width=0.21\columnwidth]{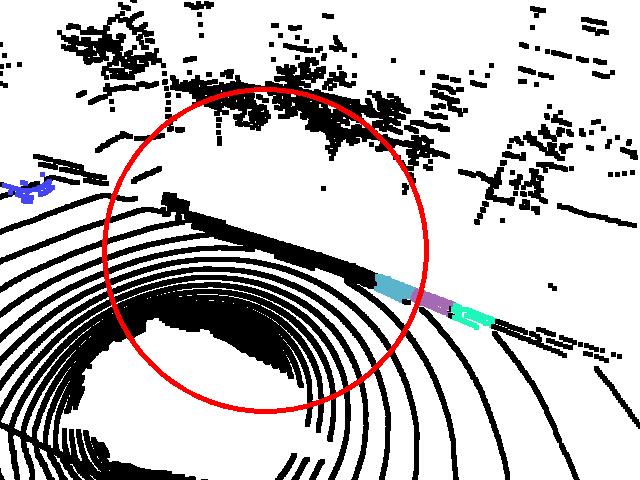}}
        \end{tabular} &
        \begin{tabular}{@{}c@{}}
            \fbox{\includegraphics[width=0.21\columnwidth]{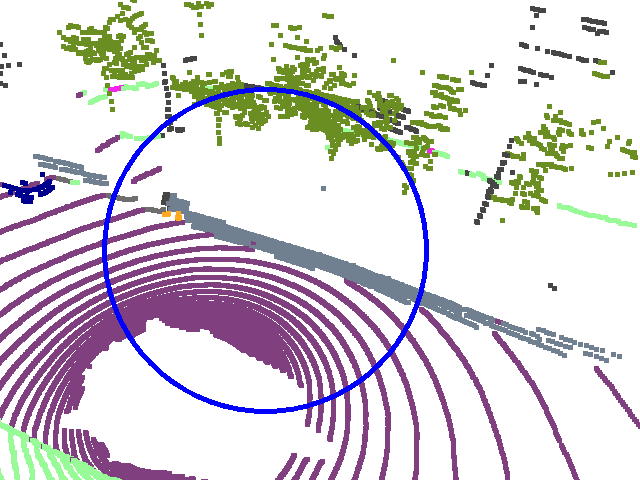}} \\[-2pt]
            \fbox{\includegraphics[width=0.21\columnwidth]{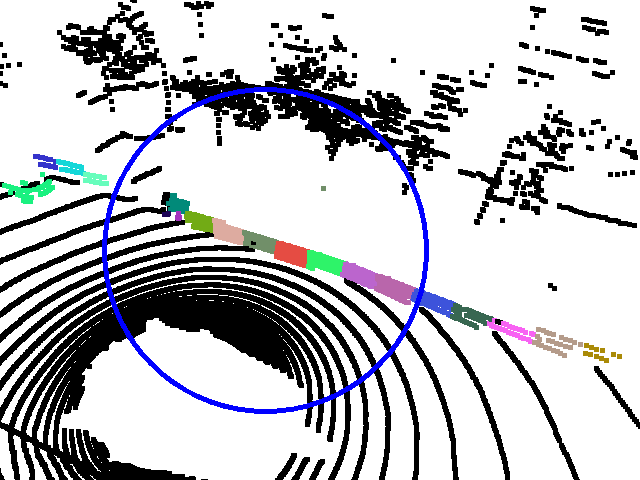}}
        \end{tabular} \\[2pt]
    \end{tabular}

    \vspace{1pt}

    \begin{tabular}{@{}m{0.3cm}cccc@{}}
        \rotatebox{90}{\scriptsize (b) Calibration Drift ($5^{\circ}$)} &
        \begin{tabular}{@{}c@{}}
            \includegraphics[width=0.22\columnwidth, height=0.11\columnwidth, keepaspectratio=false]{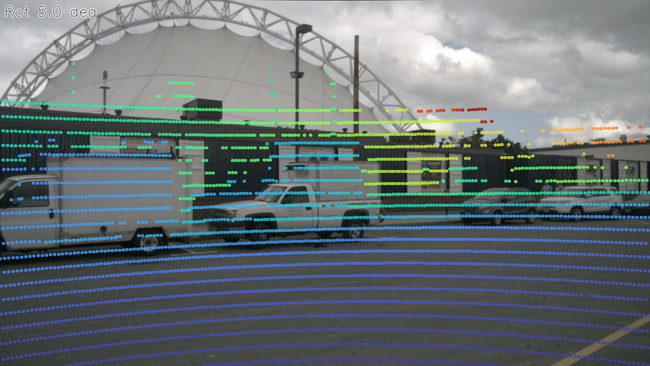} \\[1pt]
            \includegraphics[width=0.22\columnwidth, height=0.11\columnwidth, keepaspectratio=false]{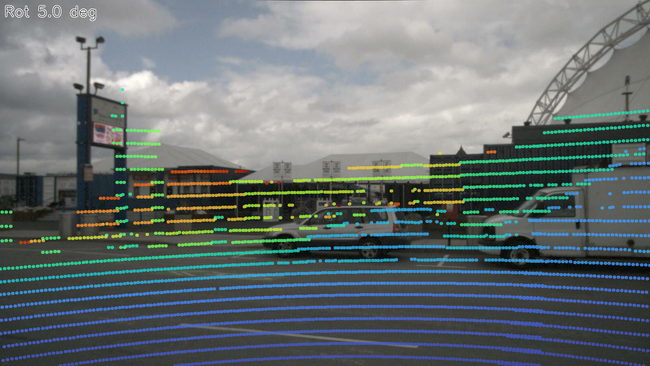}
        \end{tabular} &
        \begin{tabular}{@{}c@{}}
            \fbox{\includegraphics[width=0.21\columnwidth]{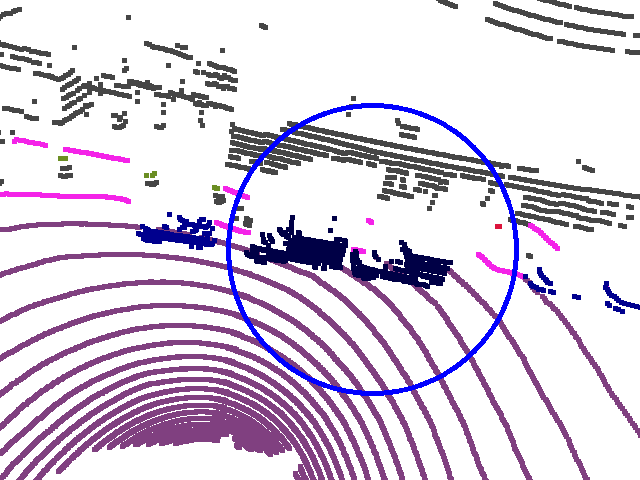}} \\[-2pt]
            \fbox{\includegraphics[width=0.21\columnwidth]{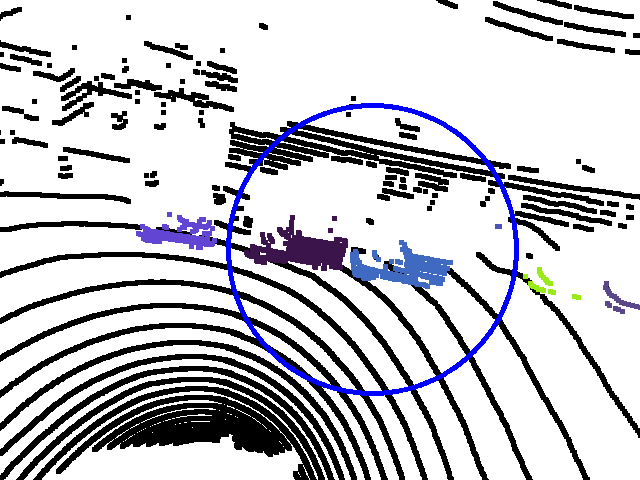}}
        \end{tabular} &
        \begin{tabular}{@{}c@{}}
            \fbox{\includegraphics[width=0.21\columnwidth]{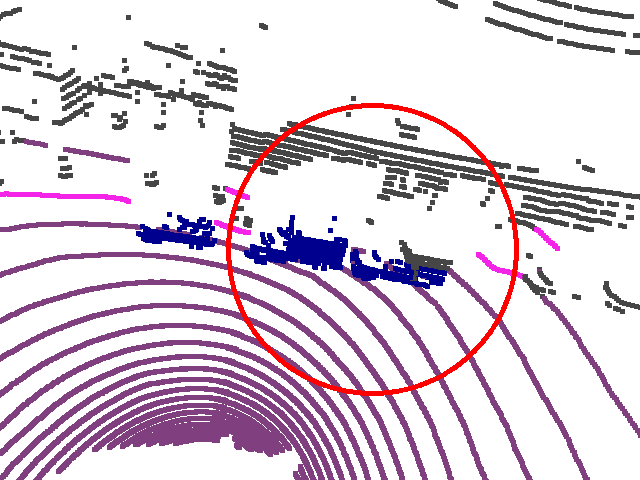}} \\[-2pt]
            \fbox{\includegraphics[width=0.21\columnwidth]{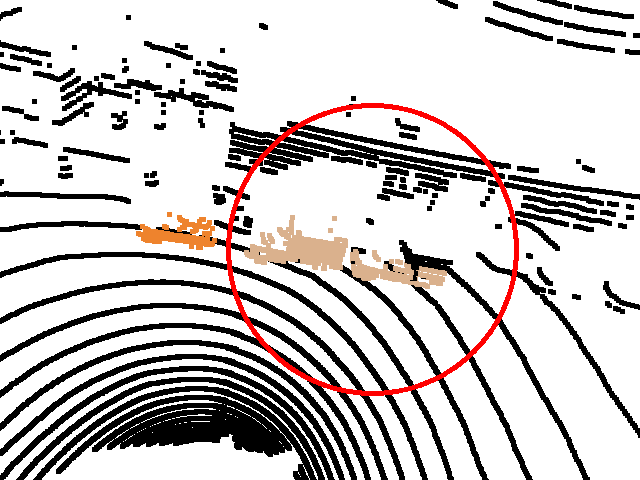}}
        \end{tabular} &
        \begin{tabular}{@{}c@{}}
            \fbox{\includegraphics[width=0.21\columnwidth]{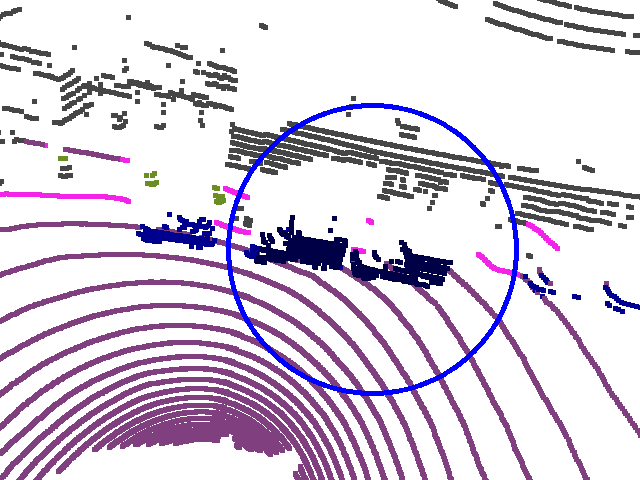}} \\[-2pt]
            \fbox{\includegraphics[width=0.21\columnwidth]{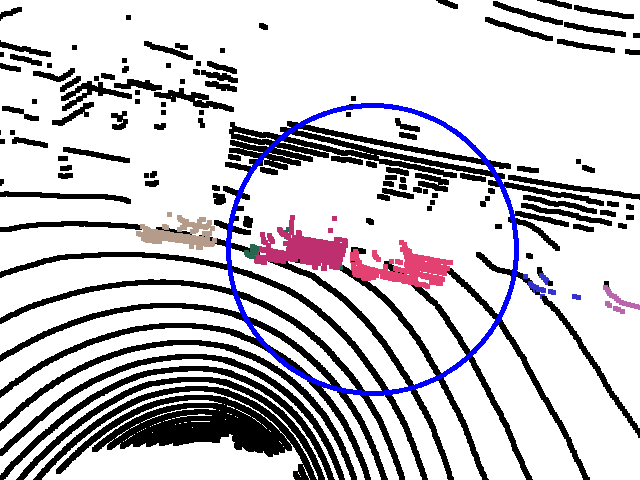}}
        \end{tabular} \\[2pt]
    \end{tabular}

    \vspace{1pt}

    \begin{tabular}{@{}m{0.3cm}cccc@{}}
        \rotatebox{90}{\scriptsize (c) Visual Domain Shift} &
        \begin{tabular}{@{}c@{}}
            \includegraphics[width=0.22\columnwidth, height=0.11\columnwidth, keepaspectratio=false]{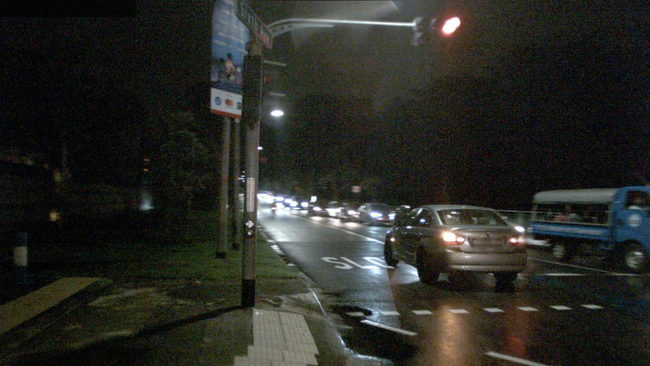} \\[1pt]
            \includegraphics[width=0.22\columnwidth, height=0.11\columnwidth, keepaspectratio=false]{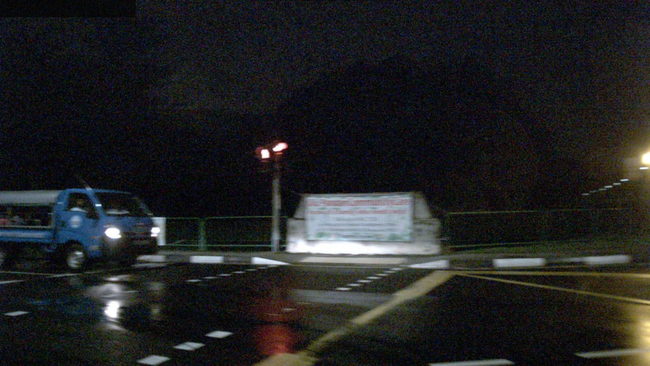}
        \end{tabular} &
        \begin{tabular}{@{}c@{}}
            \fbox{\includegraphics[width=0.21\columnwidth]{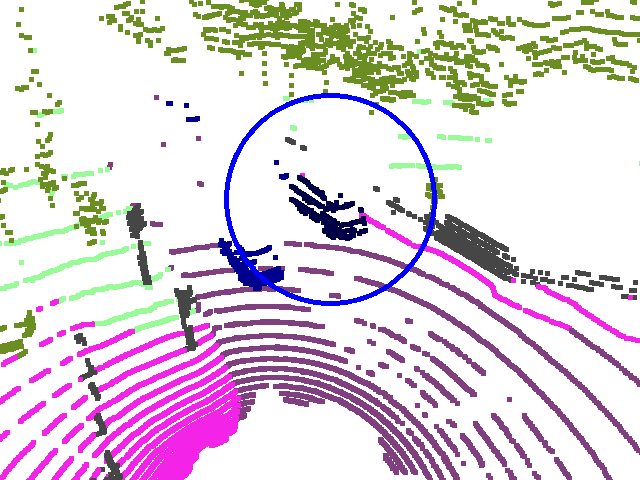}} \\[-2pt]
            \fbox{\includegraphics[width=0.21\columnwidth]{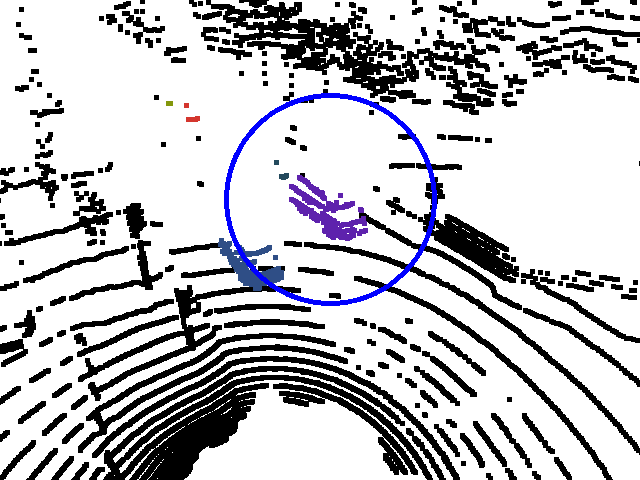}}
        \end{tabular} &
        \begin{tabular}{@{}c@{}}
            \fbox{\includegraphics[width=0.21\columnwidth]{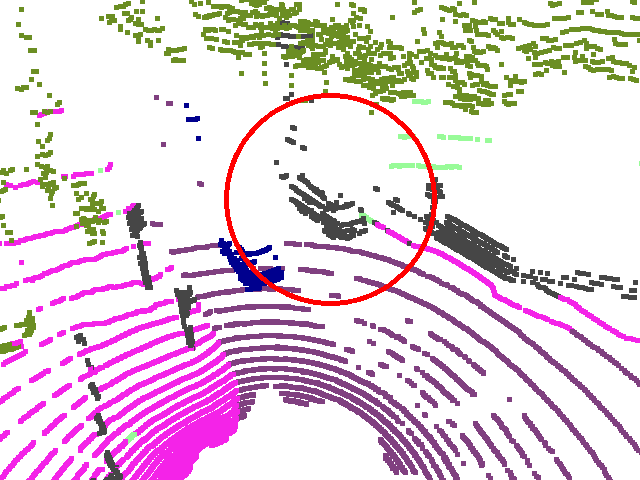}} \\[-2pt]
            \fbox{\includegraphics[width=0.21\columnwidth]{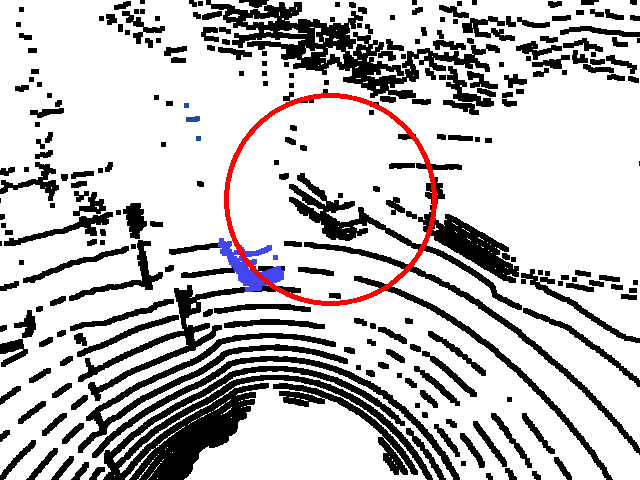}}
        \end{tabular} &
        \begin{tabular}{@{}c@{}}
            \fbox{\includegraphics[width=0.21\columnwidth]{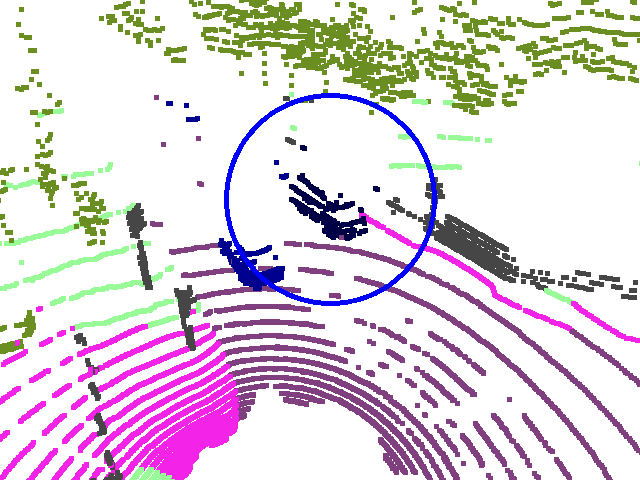}} \\[-2pt]
            \fbox{\includegraphics[width=0.21\columnwidth]{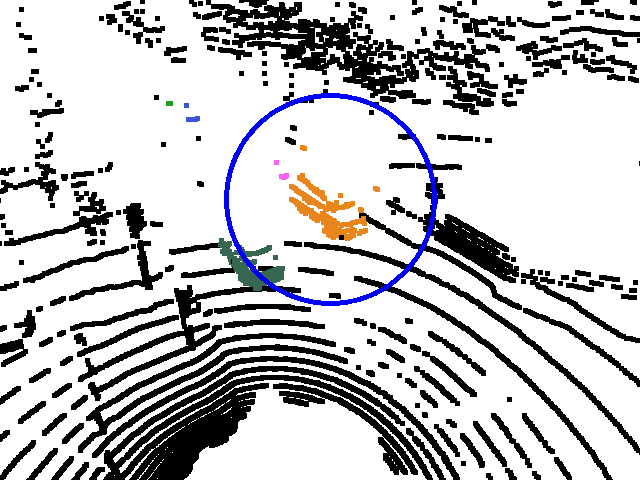}}
        \end{tabular} \\
    \end{tabular}

    \caption{Qualitative robustness comparison of 3D panoptic segmentation between UP-Fuse and the strongest baseline IAL on the Panoptic nuScenes validation set.
(a) IAL fails to resolve geometric ambiguity, misclassifying concrete barriers as man-made structures.
(b) Under calibration drift, IAL incorrectly merges two trucks into a single instance and mislabels vehicle tops as background.
(c) Nighttime domain shift leads to similar errors.
In contrast, UP-Fuse (right) suppresses unreliable visual cues and maintains correct semantic and instance predictions across all scenarios.
\textbf{\textcolor{red}{red}}: incorrect prediction, \textbf{\textcolor{blue}{blue}}: correct prediction
(Best viewed at $4\times$ zoom).}\label{fig:qual_robustness}

\end{figure}
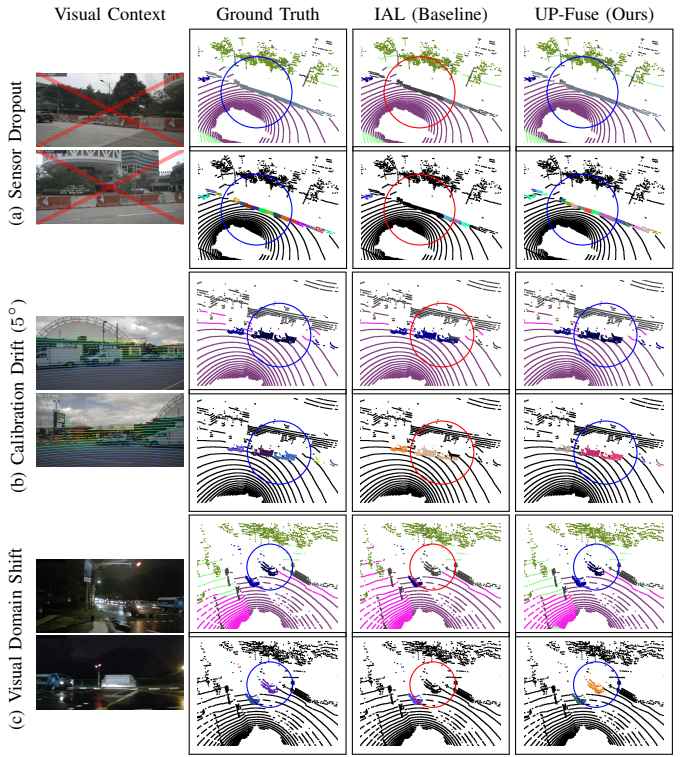

\subsection{Qualitative Evaluation}
\label{sec:qualitative}

In ~\cref{fig:qual_robustness}, we compare UP-Fuse with the strongest baseline, IAL, on the Panoptic nuScenes validation set. We show three challenging settings: sensor dropout, calibration drift, and nighttime domain shift. Under sensor dropout, IAL fails to resolve geometric ambiguity. It misclassifies concrete barriers as man-made structures. With calibration drift, IAL merges two trucks into a single instance. It also mislabels vehicle tops as background. Similar errors appear at night, where unreliable visual cues degrade both semantic and instance predictions. In contrast, UP-Fuse suppresses unreliable visual information. It preserves correct semantic labels and instance separation across all scenarios. These results show that uncertainty-aware fusion improves robustness under degraded sensing conditions.
\section{Limitations and Future Work} 

A limitation of our framework is its reliance on fixed camera parameters for view transformation. While our uncertainty module mitigates calibration drift by suppressing the visual stream, it does not explicitly correct the extrinsics. Consequently, in scenarios with severe misalignment, performance is bounded by the LiDAR-only baseline. Future work will explore joint extrinsic refinement to recover the benefits of fusion under significant sensor displacement.

Furthermore, while UP-Fuse successfully targets robustness against visual degradation, it does not currently model severe LiDAR corruption, such as ghost or missing returns. Extending this uncertainty-aware fusion to handle LiDAR degradation is an important future direction. This requires symmetric, bidirectional uncertainty modeling across both modalities. Currently, the main architectural bottleneck for this is our camera-to-range-view mapping, which explicitly depends on LiDAR depth. A more general solution would decouple this mapping from measured point clouds, for instance, through learned depth prediction.

\section{Conclusion}
In this work, we presented UP-Fuse, a robust LiDAR-Camera fusion framework for 3D panoptic segmentation built on a unified range-view representation.
UP-Fuse achieves strong performance on Panoptic nuScenes, SemanticKITTI, and our newly introduced Panoptic Waymo benchmark, while demonstrating consistent robustness across camera sensor degradation and failure.
The proposed uncertainty-guided fusion mechanism jointly models cross-modal relevance and feature reliability,
allowing the network to adaptively attenuate unreliable visual cues under degradation.
The hybrid 2D-3D transformer decoder effectively resolves spatial ambiguities inherent to range-view projections.
Together, these architectural components enable a practical balance between accuracy, efficiency, and robustness for multi-modal robotic perception.

\section*{Acknowledgments}
This research was funded by Bosch Research as part of a collaboration between Bosch Research and
the University of Freiburg on AI-based automated driving.

\newcommand{\maketitlesupplementary}{%
   \newpage
   \twocolumn[{%
    \centering
    {\LARGE\bfseries UP-Fuse: Uncertainty-guided LiDAR-Camera Fusion \\ for 3D Panoptic Segmentation\par}
    \vspace{1em}
    {\large Rohit Mohan$^{1}$, Florian Drews$^{2}$, Yakov Miron$^{2}$, Daniele Cattaneo$^{1}$, Abhinav Valada$^{1}$\par}
    \vspace{0.3em}
    {\normalsize $^{1}$University of Freiburg \quad $^{2}$Robert Bosch GmbH\par}
    \vspace{0.5em}
    {\large \textit{Supplementary Material}\par}
    \vspace{1.5em}
   }]%
}

\clearpage
\setcounter{page}{1}
\maketitlesupplementary
In this supplementary material, we present additional
details on various aspects of our work. In \cref{sec:aug_details}, we detail the full set of non-spatial image augmentations used to model aleatoric uncertainty within our UP-Fuse architecture. \cref{sec:waymo_stats} presents dataset statistics of the introduced Panoptic Waymo benchmark.
In \cref{sec:impl_details}, we provide extended implementation details, including additional architectural specifications, the inference pipeline, range-view projection parameters for each dataset, and training setups for SemanticKITTI and Panoptic Waymo baselines. \cref{sec:qual_results} presents qualitative comparisons on the Panoptic Waymo dataset. Finally, \cref{sec:real} demonstrates real-world zero-shot generalization using our in-house autonomous vehicle.

\begin{figure*}[t]
    \centering
    \begin{subfigure}{0.32\linewidth}
        \centering
        \includegraphics[width=\linewidth]{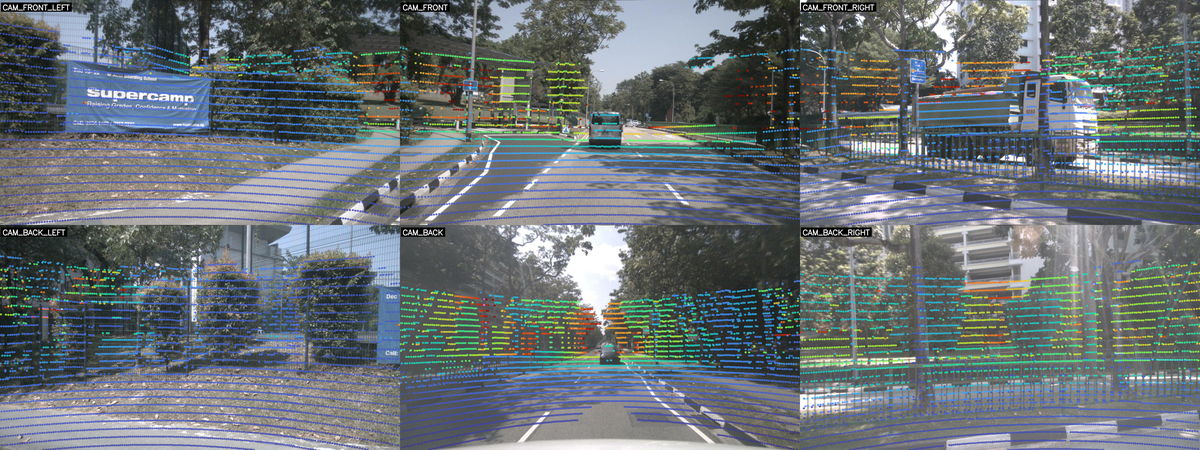}
        \caption{$0^{\circ}$}
    \end{subfigure}
    \hfill
    \begin{subfigure}{0.32\linewidth}
        \centering
        \includegraphics[width=\linewidth]{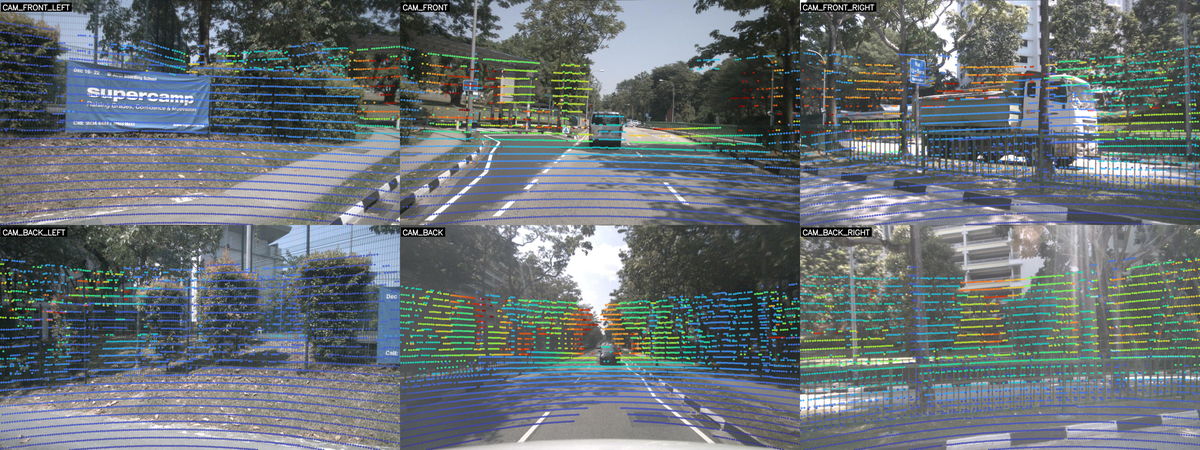}
        \caption{$1^{\circ}$}
    \end{subfigure}
    \hfill
    \begin{subfigure}{0.32\linewidth}
        \centering
        \includegraphics[width=\linewidth]{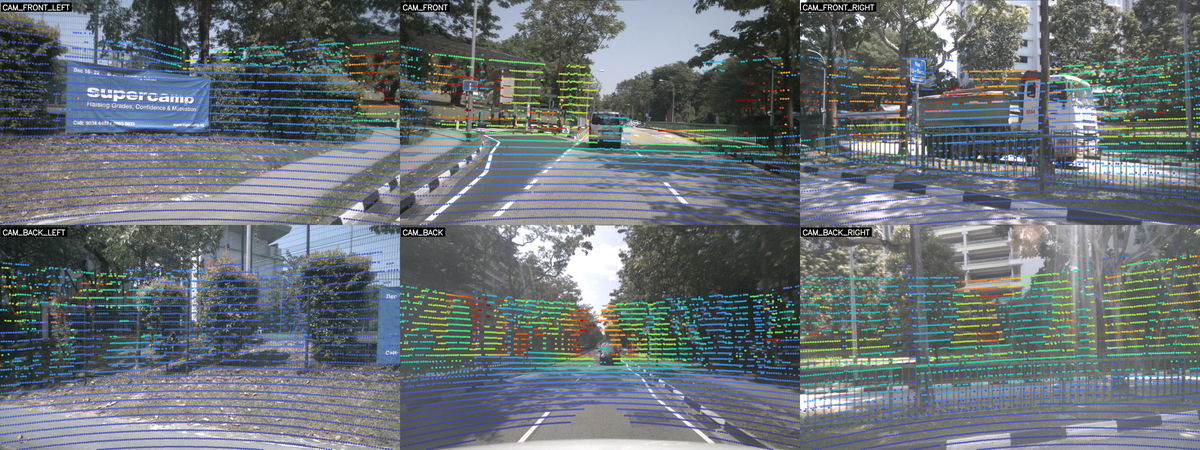}
        \caption{$2^{\circ}$}
    \end{subfigure}

    \vspace{0.6em}

    \begin{subfigure}{0.32\linewidth}
        \centering
        \includegraphics[width=\linewidth]{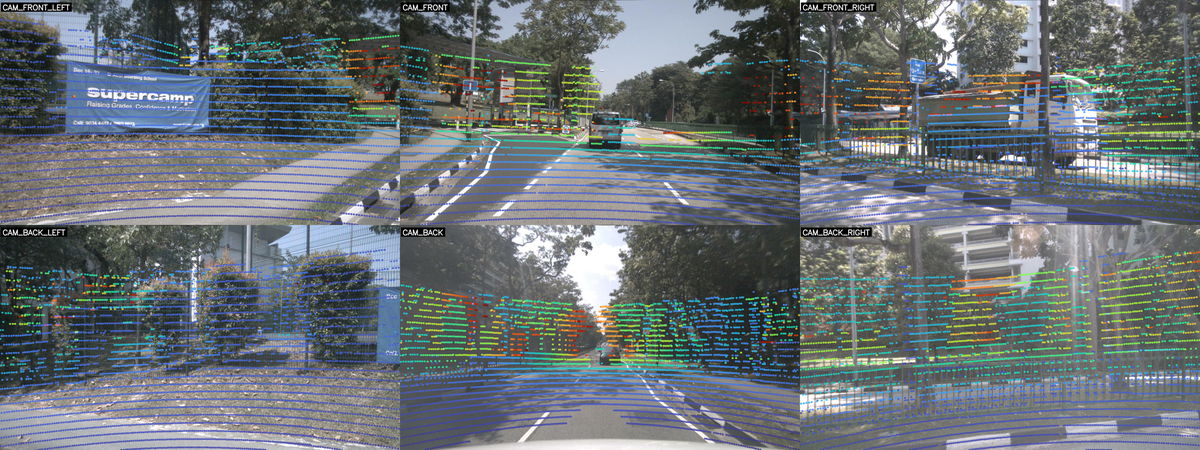}
        \caption{$3^{\circ}$}
    \end{subfigure}
    \hfill
    \begin{subfigure}{0.32\linewidth}
        \centering
        \includegraphics[width=\linewidth]{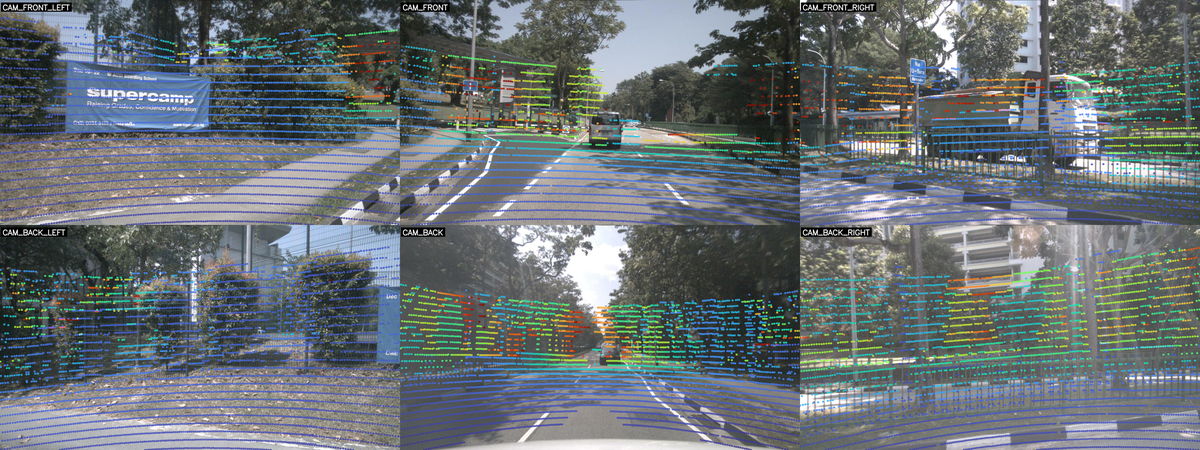}
        \caption{$4^{\circ}$}
    \end{subfigure}
    \hfill
    \begin{subfigure}{0.32\linewidth}
        \centering
        \includegraphics[width=\linewidth]{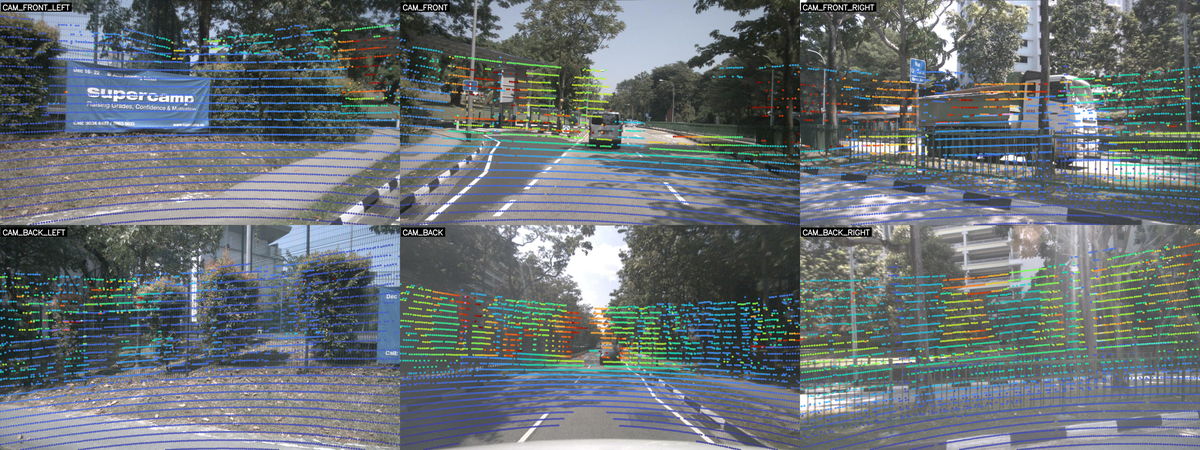}
        \caption{$5^{\circ}$}
    \end{subfigure}
    \caption{Visualization of LiDAR-to-camera projection shifts under increasing calibration drift (rotation magnitude) from $0^{\circ}$ to $5^{\circ}$, referenced in \cref{fig:plot_calib} of the main paper.}
    \label{fig:calib_projection_shift}
\end{figure*}

\section{UP-Fuse Architecture: Augmentations for Uncertainty-Aware Fusion}
\label{sec:aug_details}
To model aleatoric uncertainty as feature instability under input degradations, we apply a broad set of non-spatial augmentations to the input images. This section enumerates all augmentations used, grouped by category, along with their sampling ranges and parameters. At training time, one augmentation is sampled uniformly from the augmentation pool and applied to the input image. With a probability of $0.5$, no augmentation is applied to produce stable reference samples used for zero-uncertainty supervision.

\subsection{Photometric Augmentations}
\begin{itemize}
    \item \textit{Brightness Adjustment:} Multiplies pixel intensities by a factor sampled from $[0.7,\,1.3]$.
    \item \textit{Contrast Adjustment:} Scales the contrast around the image mean using a factor in $[0.7,\,1.3]$.
    \item \textit{Saturation Adjustment:} Modifies the saturation channel in HSV space with a factor in $[0.7,\,1.3]$.
    \item \textit{Hue Shift:} Shifts hue values by a random offset in $[-18,\,18]$ degrees.
    \item \textit{Gamma Correction:} Applies gamma correction with $\gamma \in [0.7,\,1.3]$.
    \item \textit{Color Jitter:} Sequential combination of brightness, contrast, saturation (each with probability $0.8$), and hue (probability $0.5$), using ranges:
    \[
        \text{brightness}, \text{contrast}, \text{saturation} \sim [0.6,\,1.4],
    \]
    \[
     \quad \text{hue} \sim [-18,\,18].
    \]
\end{itemize}

\subsection{Sensor and Noise Simulations}
\begin{itemize}
    \item \textit{Gaussian Noise:} Adds noise with standard deviation $\sigma \sim [5,\,25]$.
    \item \textit{Poisson Noise:} Photon noise based on image intensity distribution.
    \item \textit{Speckle Noise:} Adds multiplicative noise with scale factor $\sim [0.1,\,0.3]$.
    \item \textit{JPEG Compression:} Re-encodes the image with JPEG quality in $[40,\,95]$.
    \item \textit{Blur (Gaussian / Motion):}
    \begin{itemize}
        \item Gaussian blur with kernel size $\{3,5,7\}$.
        \item Motion blur generated using a spatially uniform linear motion field~\cite{luz2024amodal} corresponding to a blur kernel of size $[5,15]$ and random angle $[0^\circ,180^\circ]$.
    \end{itemize}
    \item \textit{Exposure Adjustment:} Exposure scaling factor sampled from $[0.5,\,1.8]$.
    \item \textit{ISO Noise Simulation:} ISO gain factor $\sim[1.0,\,2.5]$, with additive noise std. in $[10,\,30] \times \text{ISO factor}$.
\end{itemize}

\subsection{Environmental and Domain Variations}
\begin{itemize}
    \item \textit{Fog/Haze:} Blends the image with a bright fog color using intensity $\alpha \sim [0.3,\,0.7]$.
    \item \textit{Rain Streaks:} Draws $100$--$300$ streaks with random length ($5$--$20$ px) and angle ($[-15^\circ,15^\circ]$).
    \item \textit{Shadows:} Inserts $1$--$4$ random polygonal shadow regions with intensity scaling $\sim [0.3,\,0.7]$.
    \item \textit{Color Temperature Shift:} Warm/cool shift with $\Delta \sim [-50,\,50]$ applied to R/B channels.
    \item \textit{Vignette:} Darkening applied radially with intensity $\sim [0.3,\,0.7]$.
    \item \textit{White Balance Shift:} Channel-wise scaling with factors in $[0.8,\,1.2]$ for R, G, B.
    \item \textit{Histogram Matching Across Domains}:
    To induce domain shift, we perform color histogram matching against randomly sampled reference images from:
\begin{itemize}
    \item COCO~\cite{lin2014microsoft}
    \item Cityscapes~\cite{cordts2016cityscapes}
    \item Dark Zurich~\cite{sakaridis2019guided}
\end{itemize}
Histogram matching~\cite{mohan2024synmediverse, mohan2024panoptic} is performed independently per channel using cumulative distribution remapping.
    
\end{itemize}

\subsection{Sensor Failure Augmentations}
\begin{itemize}
    \item \textit{Random Dropout:} Replaces the full image with zeros (worst-case sensor dropout).
    \item \textit{Sensor Bloom / Glare:} Expands bright regions using Gaussian kernels of size $[15,40]$ with bloom intensity $[30,80]$.
\end{itemize}

\begin{figure*}[t]
    \centering
    \begin{subfigure}{0.55\linewidth}
        \centering
        \includegraphics[width=\linewidth]{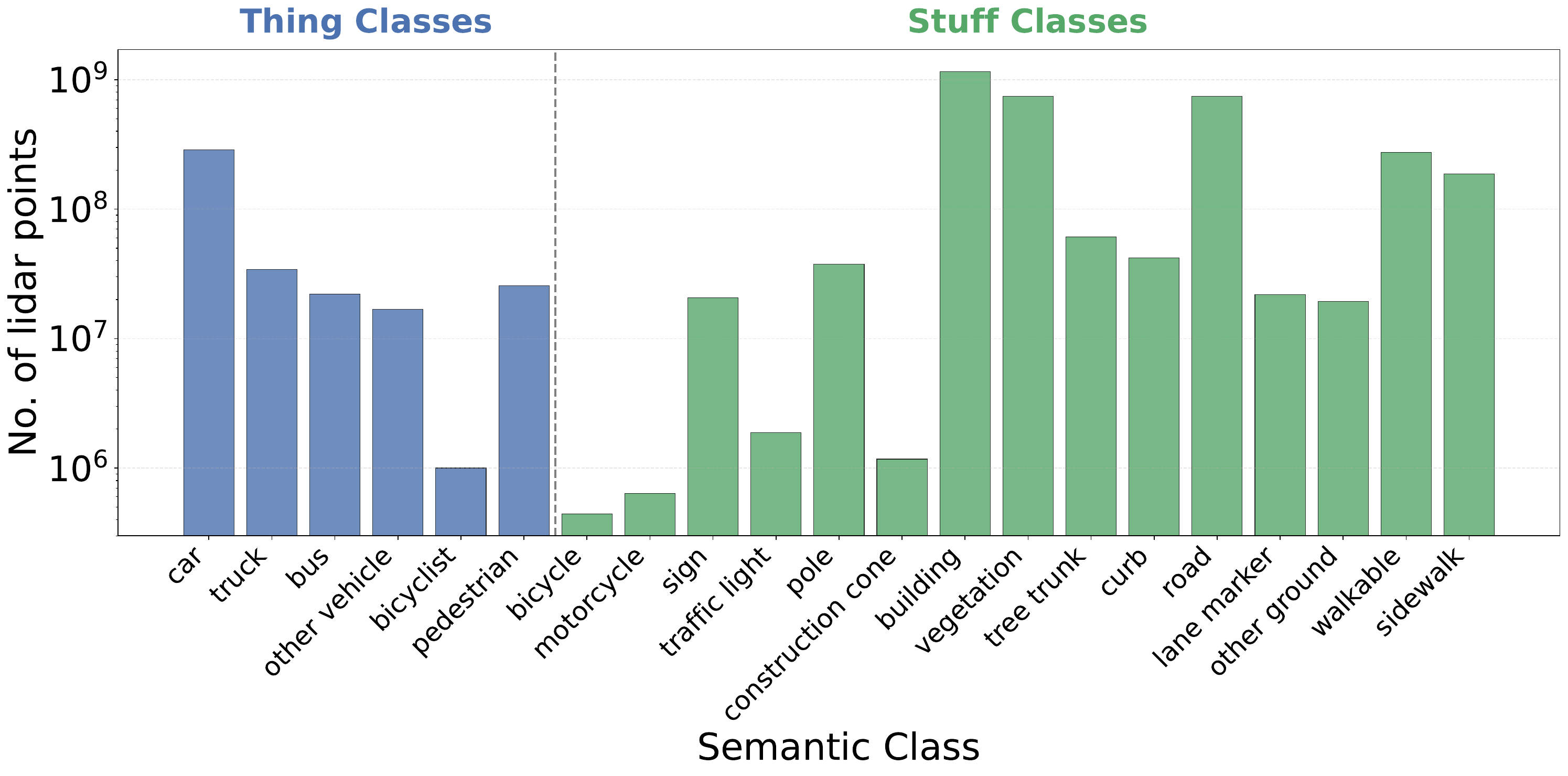}
        \caption{Number of LiDAR points for each class in Panoptic Waymo training set}
        \label{fig:lidar_points_training}
    \end{subfigure}
    \hfill
    \begin{subfigure}{0.44\linewidth}
        \centering
        \includegraphics[width=0.9\linewidth]{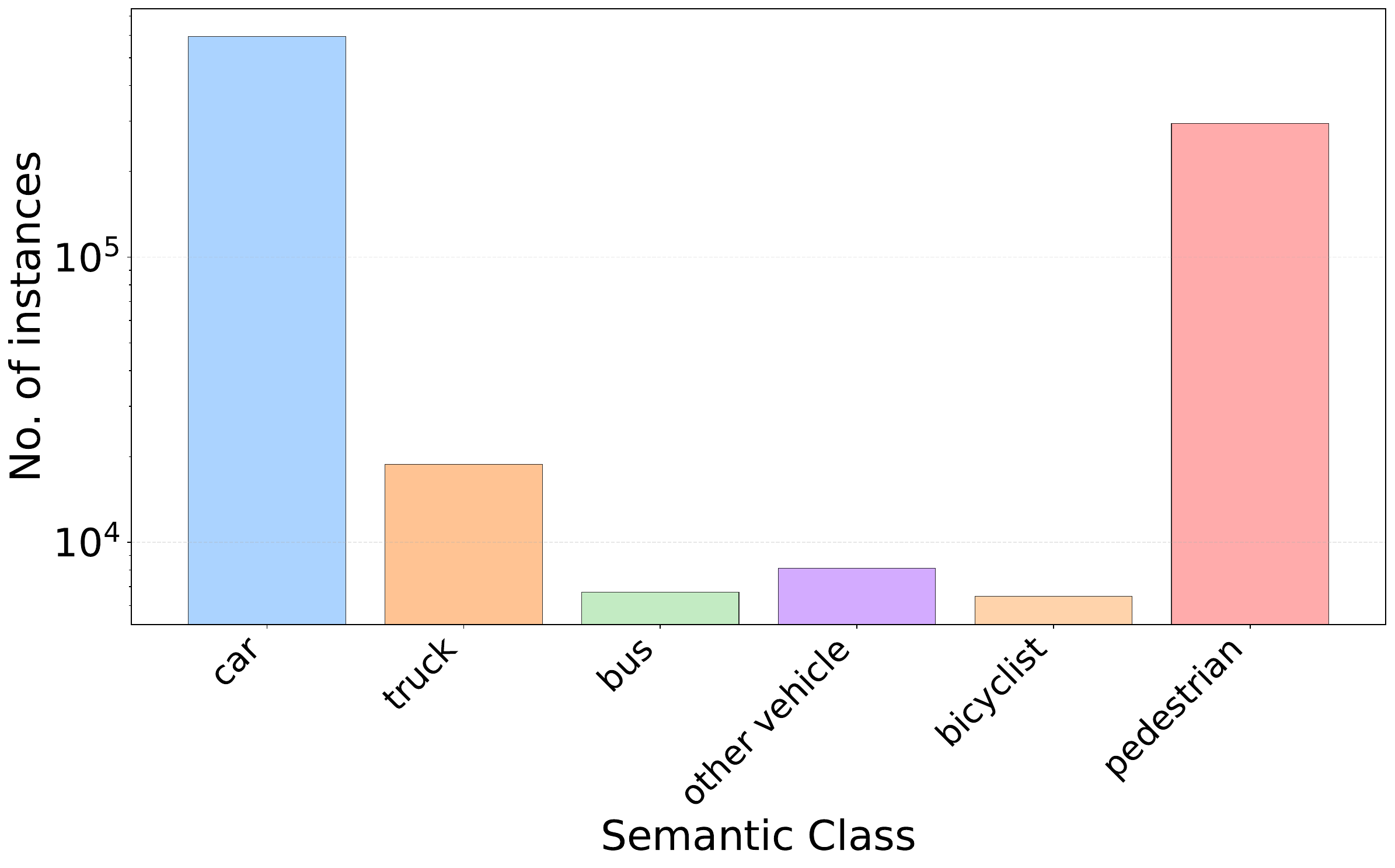}
        \caption{Number of scan-wise instances for each thing class in Panoptic Waymo training set}
        \label{fig:instances_training}
    \end{subfigure}
    \begin{subfigure}{0.55\linewidth}
        \centering
        \includegraphics[width=\linewidth]{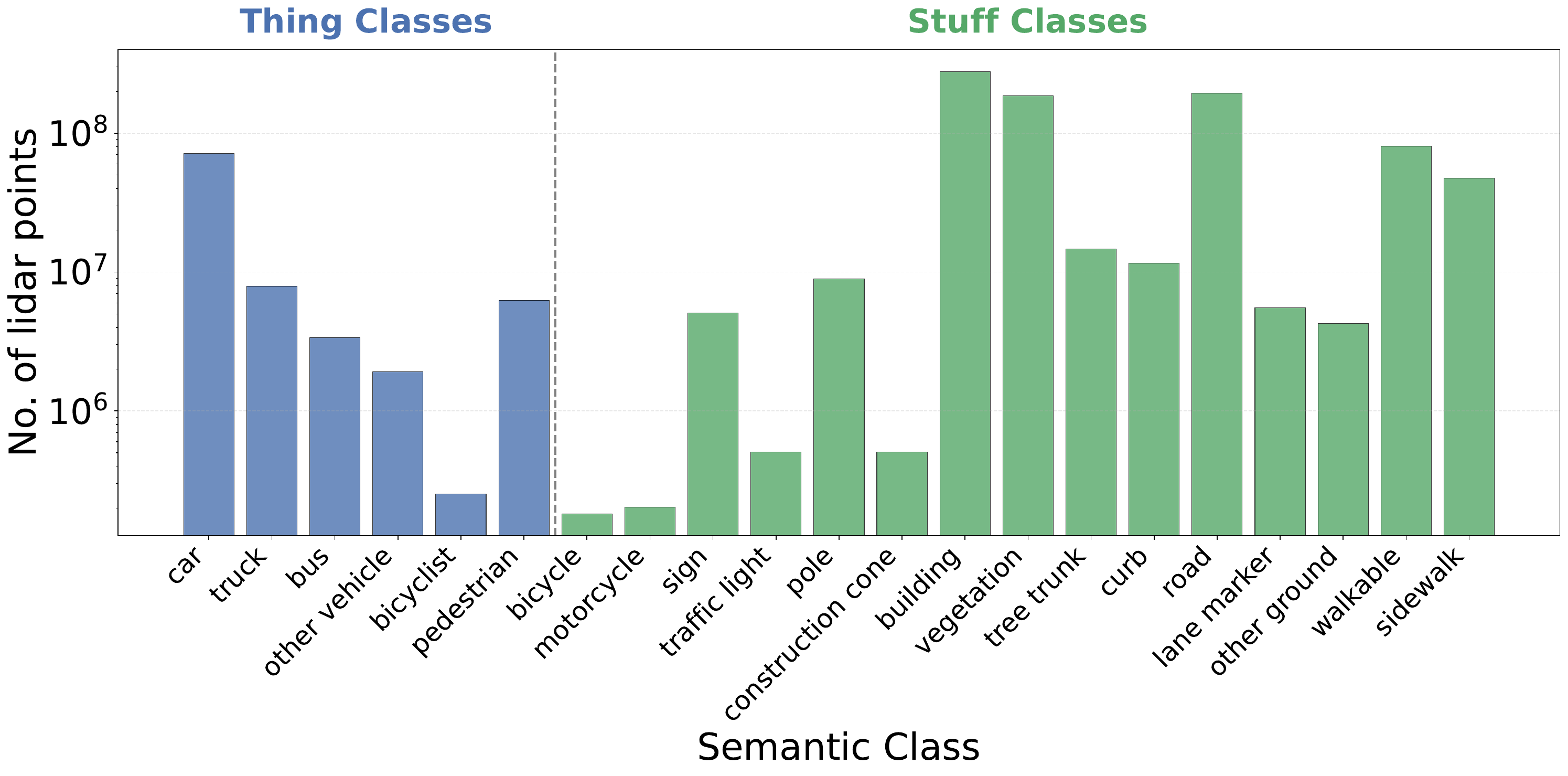}
        \caption{Number of LiDAR points for each class in the Panoptic Waymo validation set.}
        \label{fig:lidar_points_validation}
    \end{subfigure}
    \hfill
    \begin{subfigure}{0.44\linewidth}
        \centering
        \includegraphics[width=0.9\linewidth]{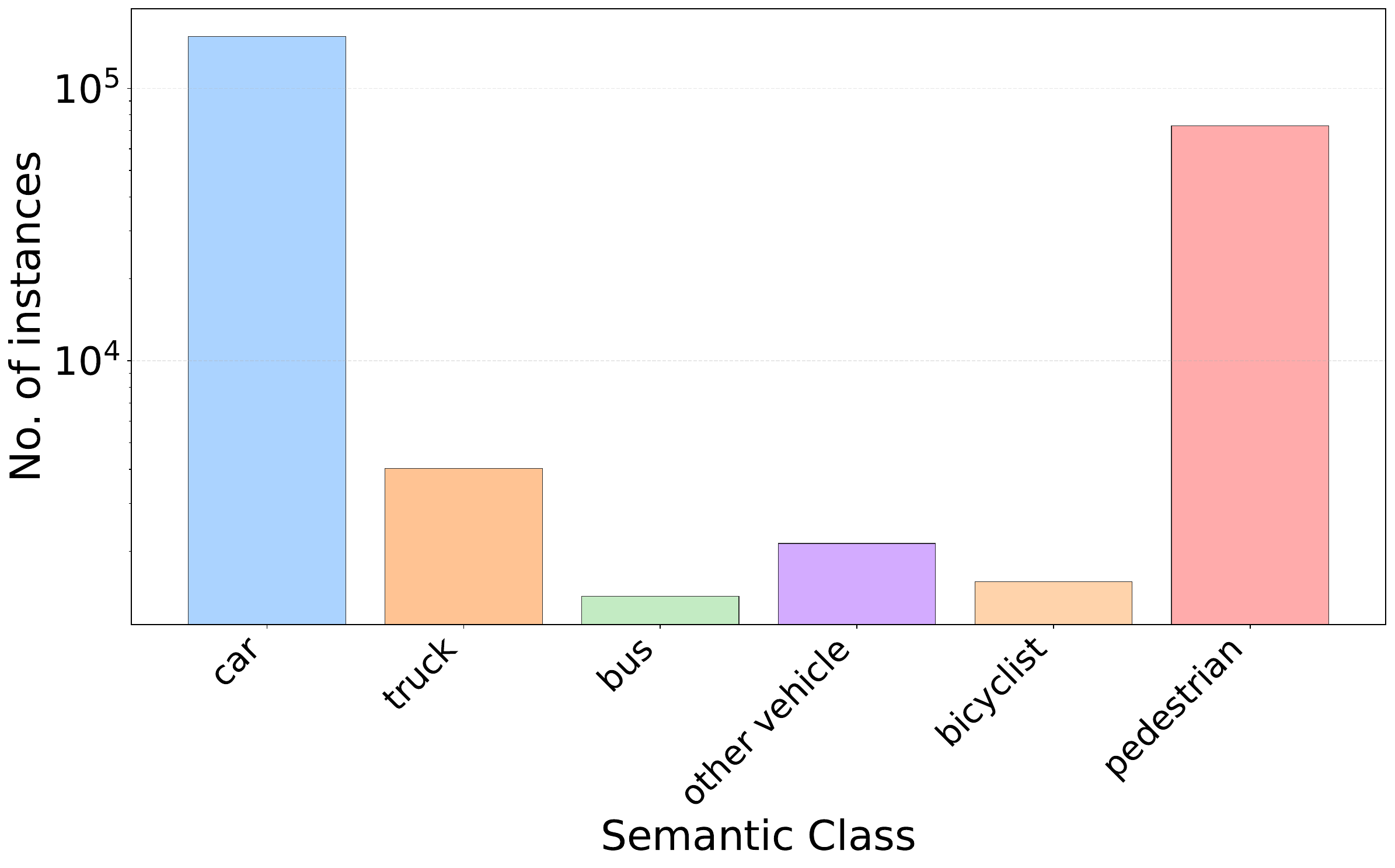}
        \caption{Number of scan-wise instances for each thing class in Panoptic Waymo validation set.}
        \label{fig:instances_validation}
    \end{subfigure}
    \caption{Dataset statistics of Panoptic Waymo.}
    \label{fig:panoptic_waymo_stastics}
\end{figure*}

\begin{figure*}
  \centering
  \includegraphics[width=0.8\linewidth]{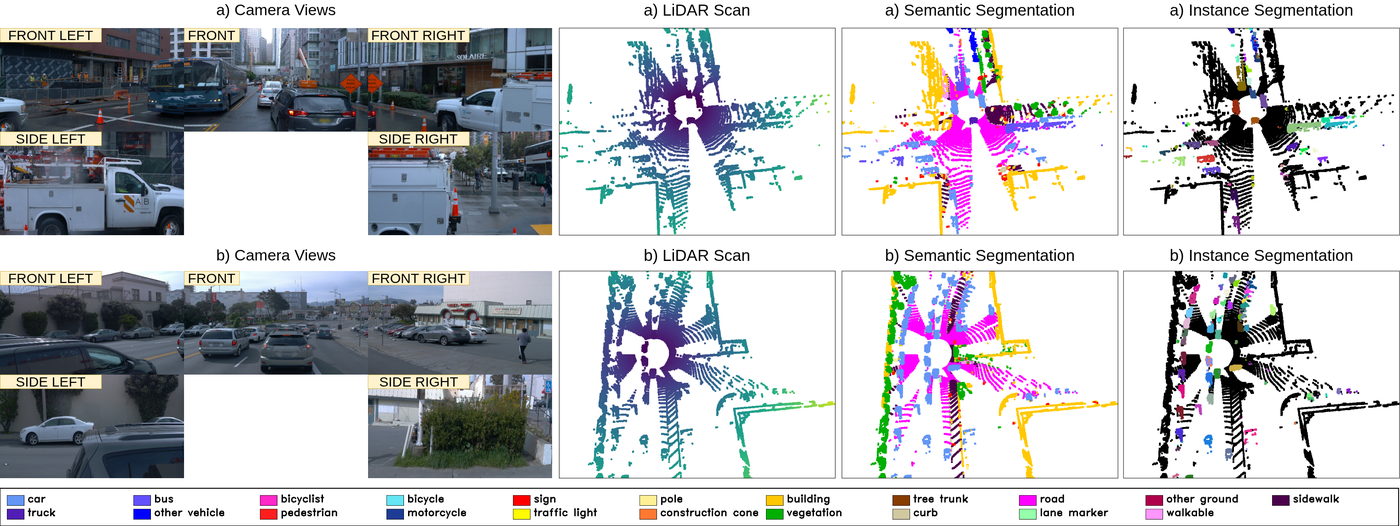}
  \caption{Examples from the Panoptic Waymo dataset showing the multi-view camera images, corresponding LiDAR scan, and their panoptic ground truth. For clarity, the semantic and instance components of the panoptic labels are visualized separately.}
  \label{fig:panoptic_waymo_gt}
\end{figure*}

\section{Statistics of the Panoptic Waymo Dataset}
\label{sec:waymo_stats}
The Waymo Open Dataset~\cite{sun2020scalability} contains 5.2B annotated LiDAR points across 22 semantic classes, offering more than five times the annotated LiDAR points of Panoptic nuScenes~\cite{fong2022panoptic}. In addition, the dataset provides temporally consistent 3D bounding boxes for vehicles, pedestrians, and two-wheeler categories. For our Panoptic Waymo setup, we merge the point-level semantic labels with the 3D bounding boxes to derive instance IDs. Each instance is defined as the set of points that (i) lie inside a given 3D bounding box and (ii) share the same semantic class as the box. In particular, points within boxes labeled as \emph{vehicle} are assigned instance labels corresponding to \emph{car}, \emph{truck}, \emph{bus}, or \emph{other vehicle}. Pedestrian boxes map to the \emph{pedestrian} class, and \emph{two-wheeler} boxes map to the \emph{bicyclist} and \emph{motorcyclist} classes. Since the dataset contains only 28 motorcycle instances in the training set and none in the validation set, we remove this class from our Panoptic Waymo benchmark. 

\cref{fig:lidar_points_training} and \cref{fig:lidar_points_validation} show the number of LiDAR points per semantic class for the training and validation splits, respectively. \cref{fig:instances_training} and \cref{fig:instances_validation} further illustrate the number of instances for each of the \textit{thing} classes. Among the \textit{stuff} classes, the most frequent categories are \emph{road}, \emph{building}, and \emph{tree trunk}, whereas among the \textit{thing} classes, dynamic objects such as \emph{car} and \emph{pedestrian} dominate in frequency. The training split contains a maximum of 219 instances per scan, with an average of 39 instances. In comparison, the validation split has a maximum of 178 instances per scan, also averaging 39 instances per scan. \cref{fig:panoptic_waymo_gt} presents examples of camera views, the corresponding LiDAR scans, and their panoptic labels visualized separately for semantic and instance components for clarity.

\section{Implementation details}
\label{sec:impl_details}
This section provides additional implementation details. We first describe extended architectural details of our fusion module and hybrid decoder, as well as our inference scheme. We then outline the range-view projection parameters that we use for each dataset to ensure consistent LiDAR encoding. Finally, we provide the training protocols used for baseline comparisons on Panoptic Waymo.

\subsection{Additional Architecture Details and Inference}
In \cref{sec:implementation_details}, we describe our use of Swin-B~\cite{liu2021swin} backbones for both LiDAR and camera, as well as the pre-training strategy for the camera encoder. The LiDAR and camera encoders output multi-scale features $\mathbf{F}_{L,s}$ and $\mathbf{F}_{C,s}$ at resolutions $s \in \{4, 8, 16, 32\}$, with corresponding channel dimensions $D_s \in \{128,\,256,\,512,\,1024\}$. For each scale, the predicted instability $\mathbf{d}_{\text{pred},s}$ in our uncertainty-aware fusion module is produced by a lightweight 3-layer MLP ($\mathcal{U}_{\theta,s}$). This MLP first expands the feature dimension from $D_s$ to $2D_s$ using a linear layer with ReLU, processes it with a second linear layer that preserves the $2D_s$ dimension and applies another ReLU, and finally projects the result to a single channel using a third linear layer. The deformable attention layers in our fusion module (\cref{sec:fusion}) use an embedding dimension of $D_s$ at each scale. The 3D-aware mask head in our 2D-3D hybrid panoptic decoder aggregates features from $K = 5$ neighbors using a 2-layer $1\times1$ convolutional MLP. The first convolution expands the concatenated neighbor features from $K D_o$ to $2D_o$ channels, followed by a ReLU activation, and the second convolution reduces the dimension back to $D_o = 256$, producing the final per-query aggregated feature. The remaining pixel decoder and transformer decoder settings follow \cite{cheng2022masked}.

\noindent\textbf{Inference:}
At inference time, each panoptic query predicts both a class probability and a point-level mask. We compute confidence-modulated mask scores by multiplying the class confidence with the predicted mask, and each point is assigned to the query with the highest such score to obtain the panoptic segmentation. Stuff classes are merged into one region per class, whereas thing classes produce distinct instance IDs.

\subsection{Range-View Projection Parameters}

For all experiments, we apply dataset-specific field-of-view settings when projecting LiDAR points into the range view. These parameters follow the definitions provided in \cref{sec:lidar_projection_encoding}, where $f_{\text{up}}$ and $f_{\text{down}}$ denote the vertical angular limits above and below the horizontal plane, and $f_{\text{left}}$ and $f_{\text{right}}$ specify the horizontal angular limits to the left and right of the forward axis.
We adopt the following values for each dataset:

\begin{itemize}
    \item \textit{Panoptic nuScenes:} 
    $f_{\text{up}} = 10^\circ$, 
    $f_{\text{down}} = -30^\circ$, \\
    $f_{\text{left}} = -180^\circ$, 
    $f_{\text{right}} = 180^\circ$.

    \item \textit{SemanticKITTI:} 
    $f_{\text{up}} = 10^\circ$, 
    $f_{\text{down}} = -30^\circ$, \\
    $f_{\text{left}} = -180^\circ$, 
    $f_{\text{right}} = 180^\circ$.

    \item \textit{Panoptic Waymo:} 
    $f_{\text{up}} = 2.4^\circ$, 
    $f_{\text{down}} = -17.6^\circ$, \\
    $f_{\text{left}} = -180^\circ$, 
    $f_{\text{right}} = 180^\circ$.
\end{itemize}
These settings ensure that the spherical range-view projection reflects the native sensor characteristics of each dataset.

\begin{figure*}
  \centering
  \includegraphics[width=0.8\linewidth]{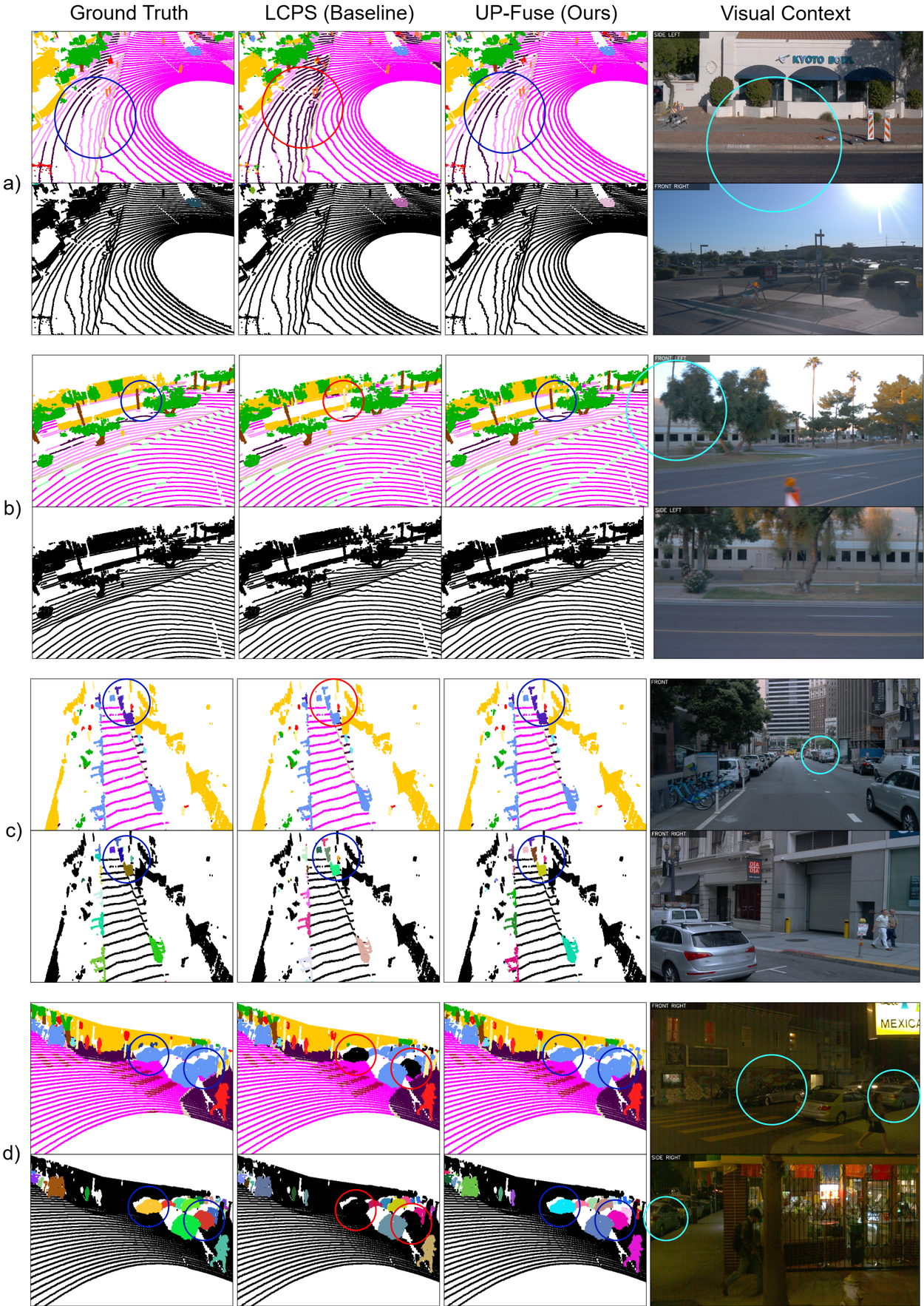}
  \caption{Qualitative 3D Panoptic Segmentation results of our proposed UP-Fuse network versus the baseline LCPS architecture, on the Panoptic Waymo val set. \textbf{\textcolor{red}{red}}: incorrect prediction, \textbf{\textcolor{blue}{blue}}: correct prediction.}
  \label{fig:qual_waymo}
\end{figure*}

\begin{figure*}
    \centering
    \setlength{\tabcolsep}{1pt}
    \renewcommand{\arraystretch}{0.1}
    \footnotesize

    \begin{tabular}{@{}>{\centering\arraybackslash}m{0.5cm}@{\hspace{2pt}}*{4}{>{\centering\arraybackslash}m{0.235\textwidth}}@{}}
        &
        \scriptsize LiDAR Point Cloud &
        \scriptsize Front View Camera &
        \scriptsize UP-Fuse Semantic Prediction &
        \scriptsize UP-Fuse Instance Prediction \\[3pt]

        \scriptsize (a) &
        \includegraphics[width=0.235\textwidth]{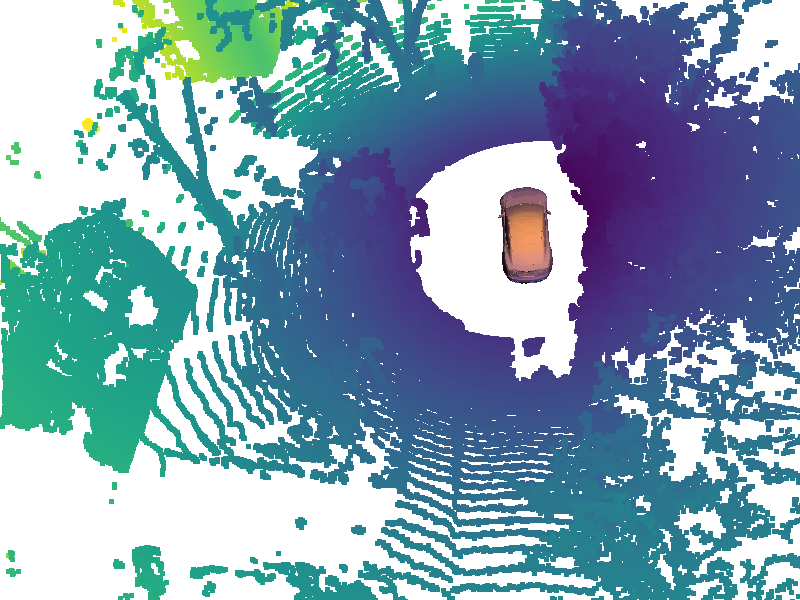} &
        \includegraphics[width=0.235\textwidth]{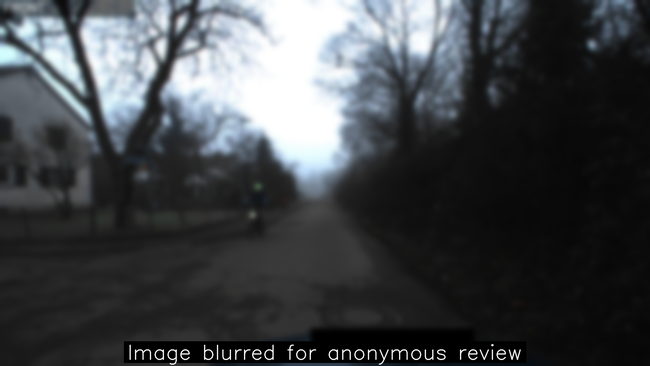} &
        \includegraphics[width=0.235\textwidth]{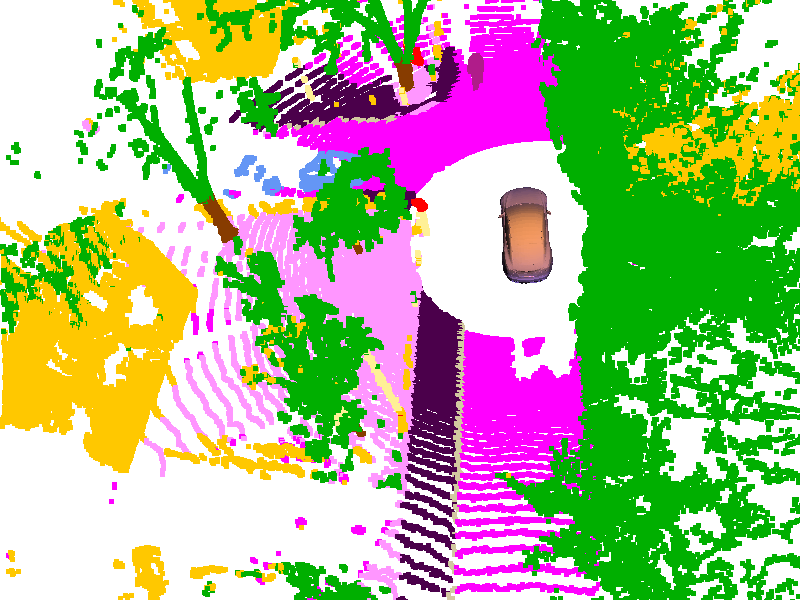} &
        \includegraphics[width=0.235\textwidth]{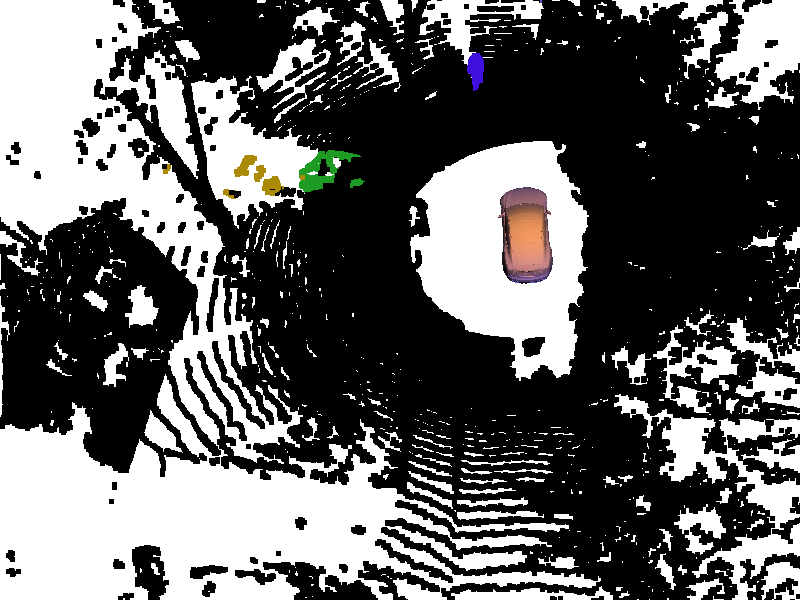} \\[2pt]

        \scriptsize (b) &
        \includegraphics[width=0.235\textwidth]{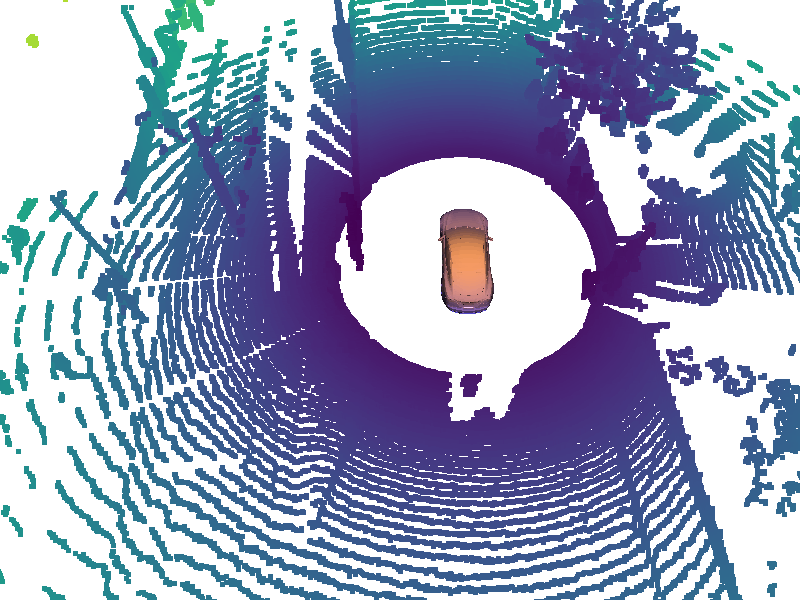} &
        \includegraphics[width=0.235\textwidth]{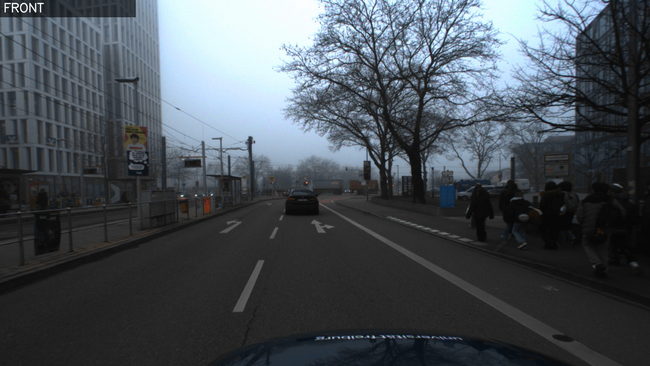} &
        \includegraphics[width=0.235\textwidth]{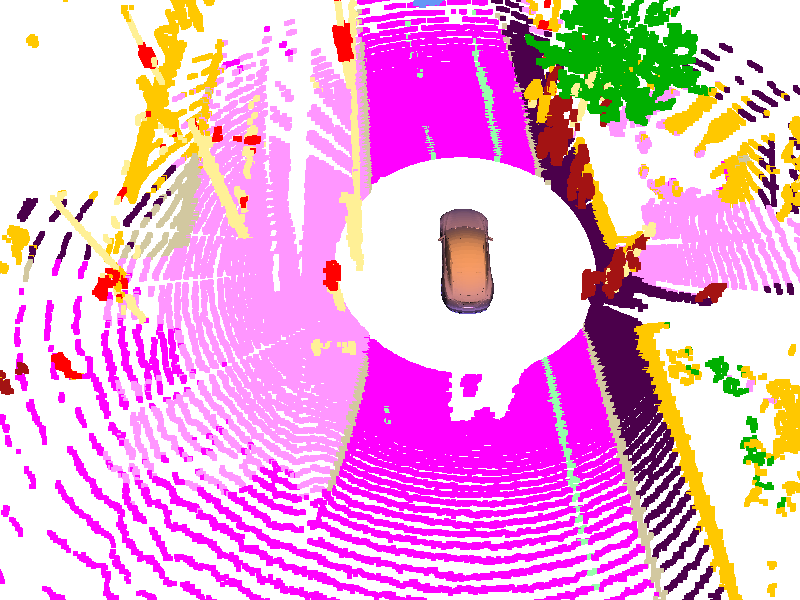} &
        \includegraphics[width=0.235\textwidth]{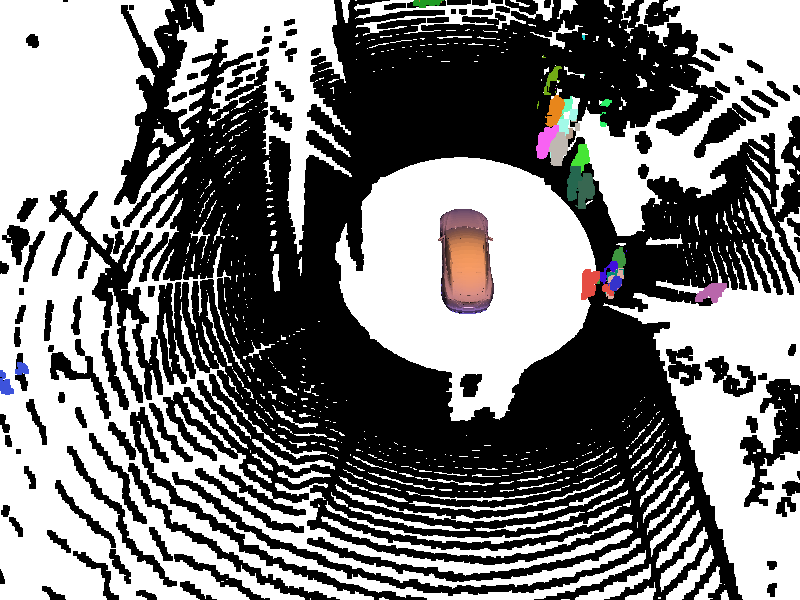} \\[2pt]

        \scriptsize (c) &
        \includegraphics[width=0.235\textwidth]{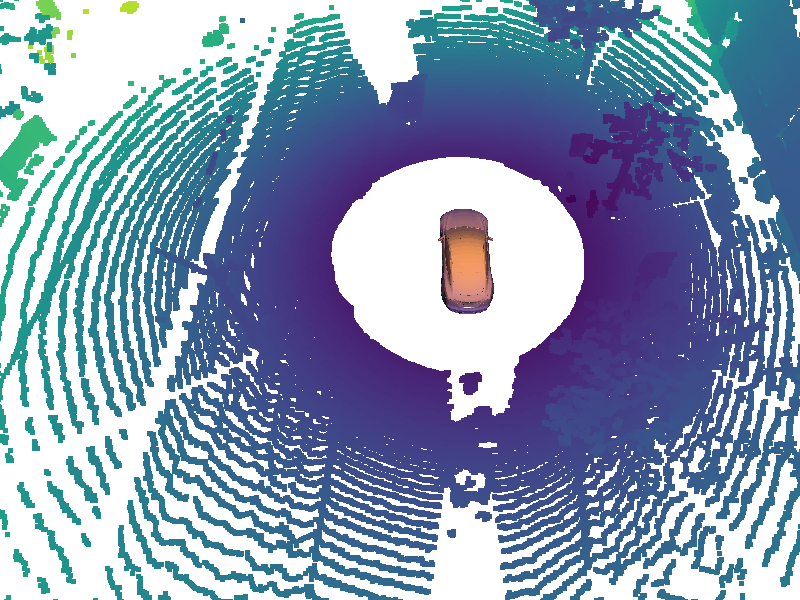} &
        \includegraphics[width=0.235\textwidth]{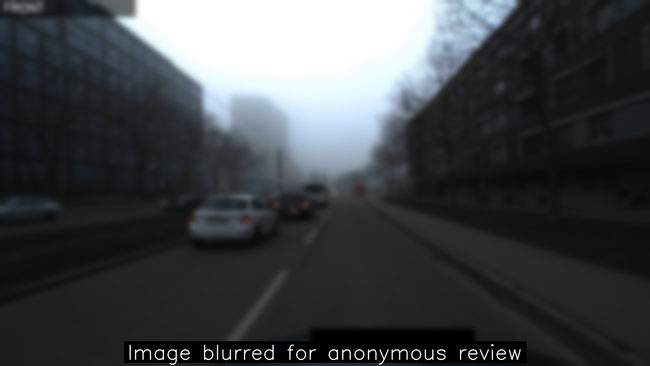} &
        \includegraphics[width=0.235\textwidth]{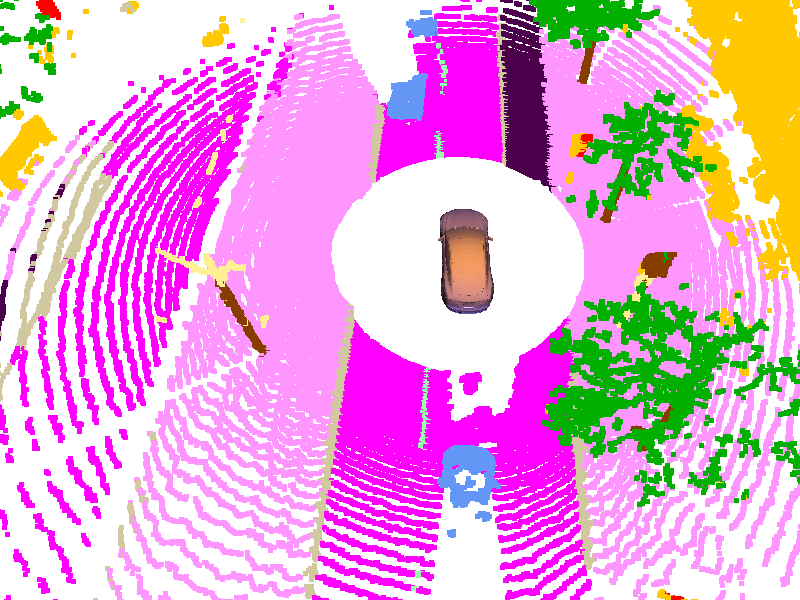} &
        \includegraphics[width=0.235\textwidth]{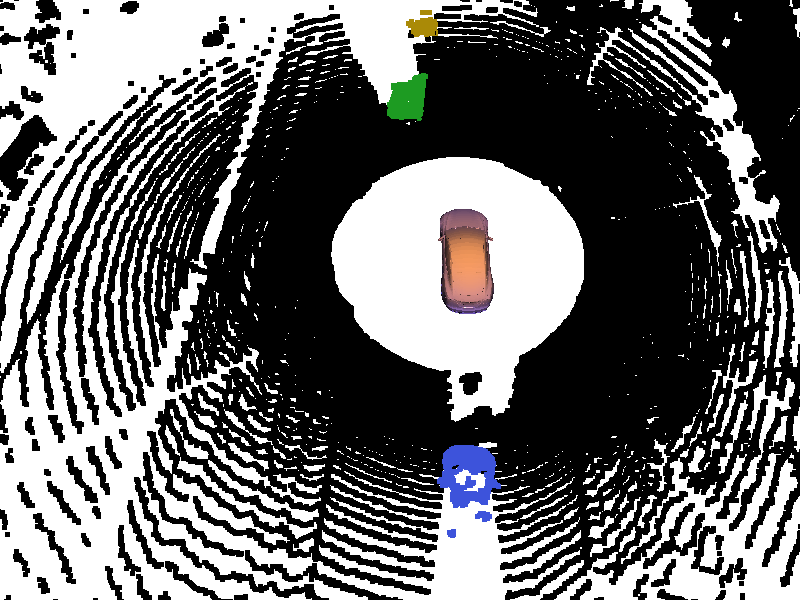} \\[2pt]

        \scriptsize (d) &
        \includegraphics[width=0.235\textwidth]{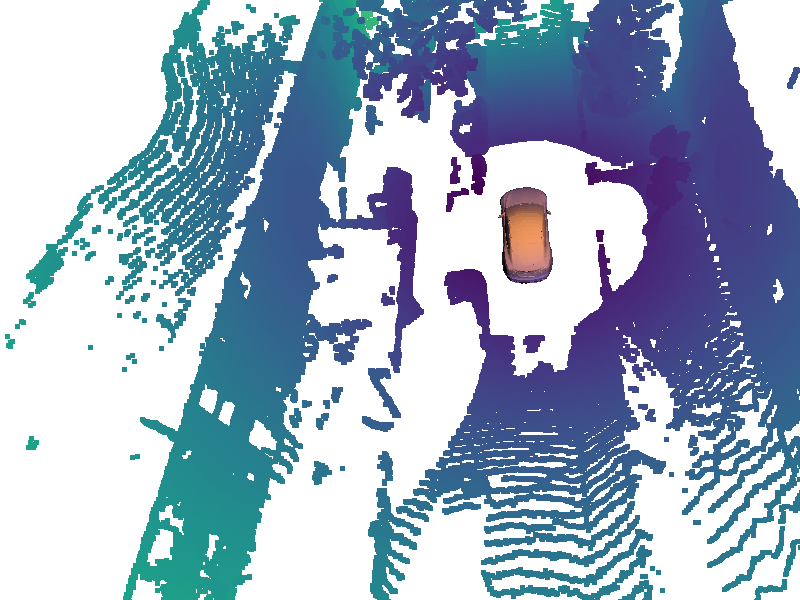} &
        \includegraphics[width=0.235\textwidth]{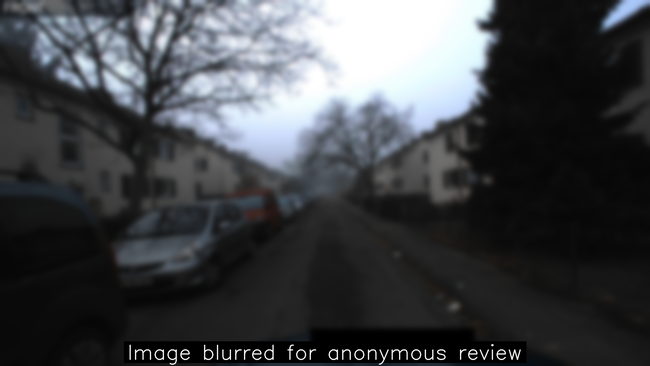} &
        \includegraphics[width=0.235\textwidth]{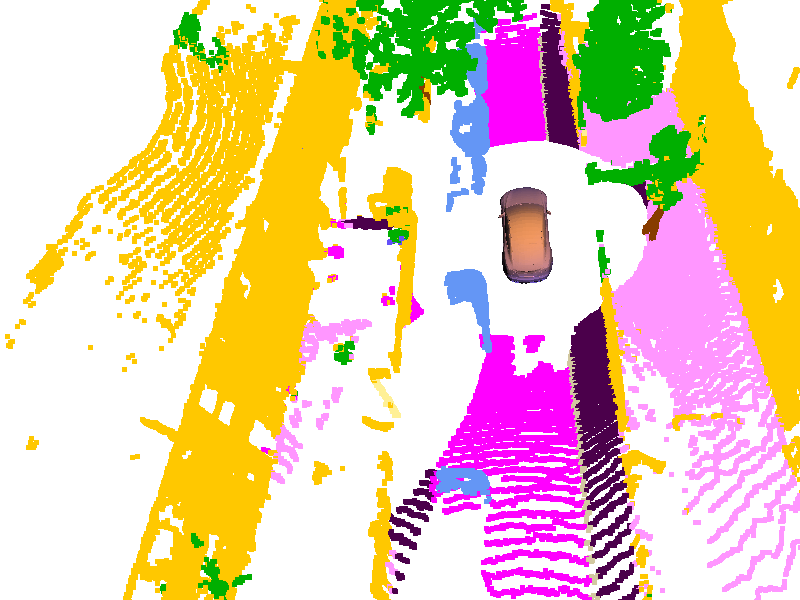} &
        \includegraphics[width=0.235\textwidth]{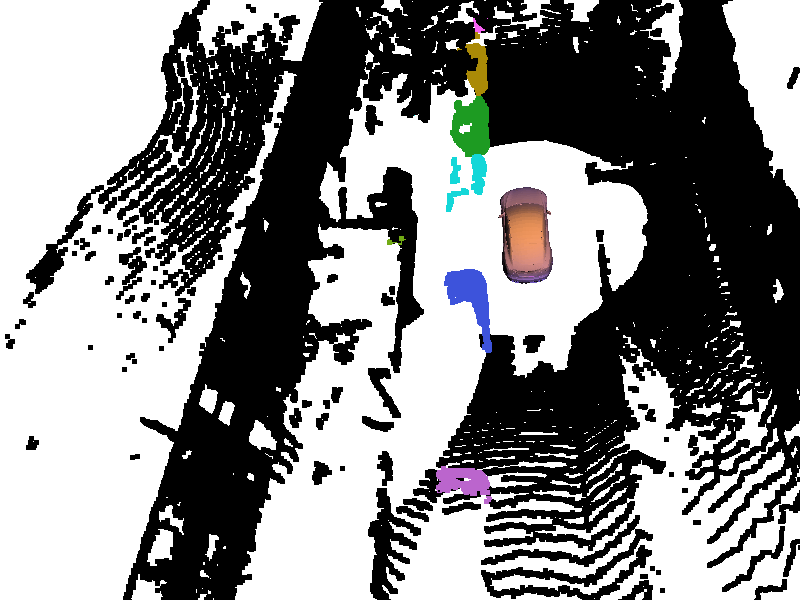} \\[2pt]

        \scriptsize (e) &
        \includegraphics[width=0.235\textwidth]{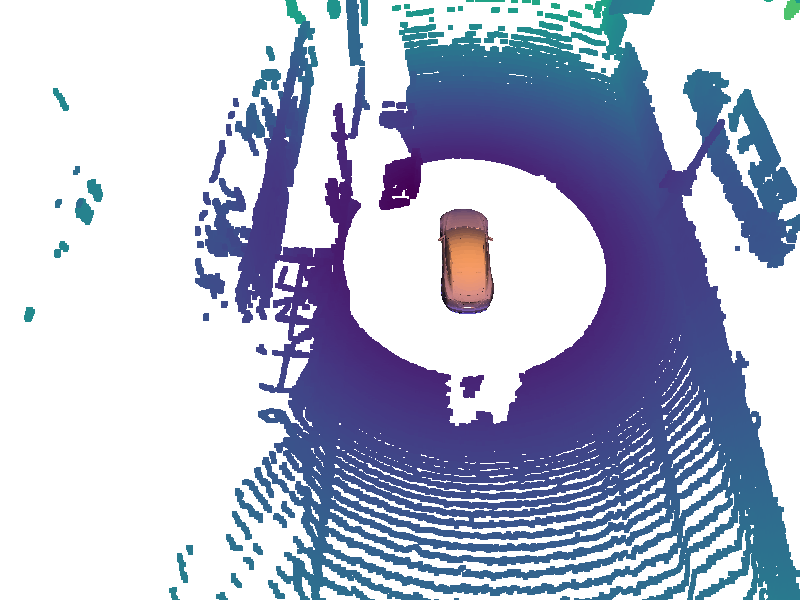} &
        \includegraphics[width=0.235\textwidth]{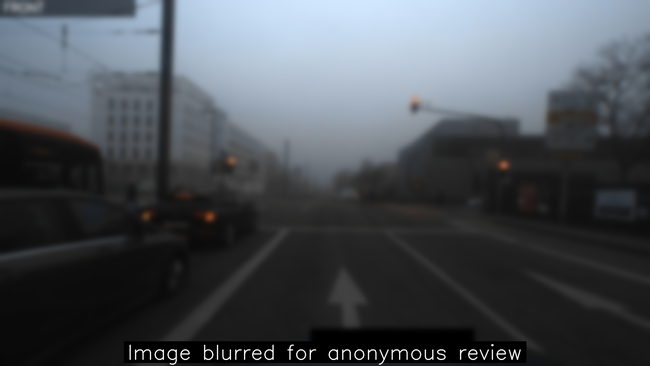} &
        \includegraphics[width=0.235\textwidth]{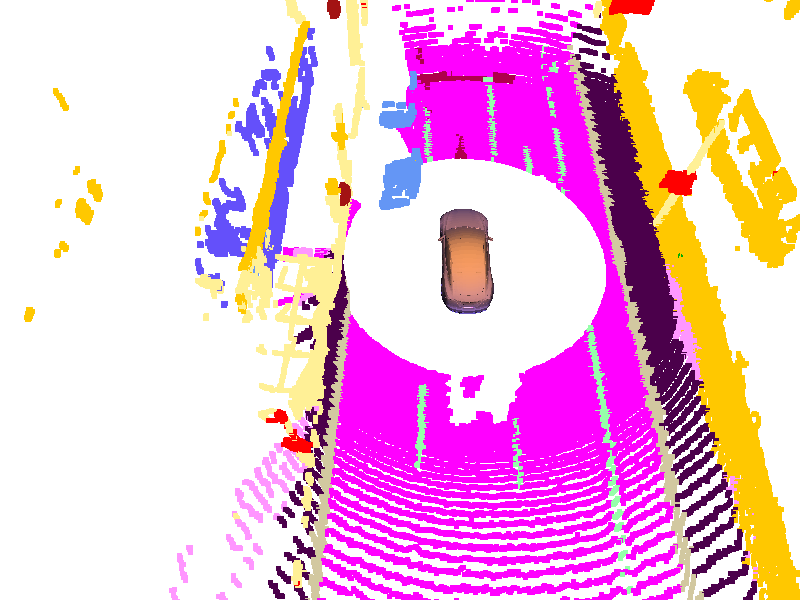} &
        \includegraphics[width=0.235\textwidth]{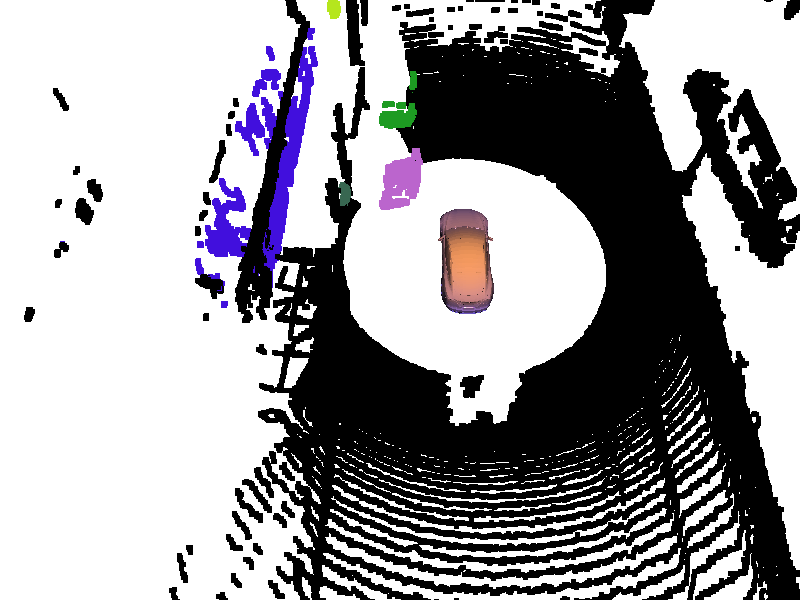}
    \end{tabular}

    \caption{Visualization of 3D panoptic segmentation predictions of UP-Fuse on real-world scenes. A model trained on the Panoptic Waymo dataset is deployed on sensor inputs from our in-house autonomous vehicle, highlighting its behavior under domain shift, differences in sensor configuration, and missing camera views compared to the original Panoptic Waymo setup.}\label{fig:real_world_qualitative}
\end{figure*}

\begin{figure}
    \centering
        \includegraphics[width=0.5\linewidth]{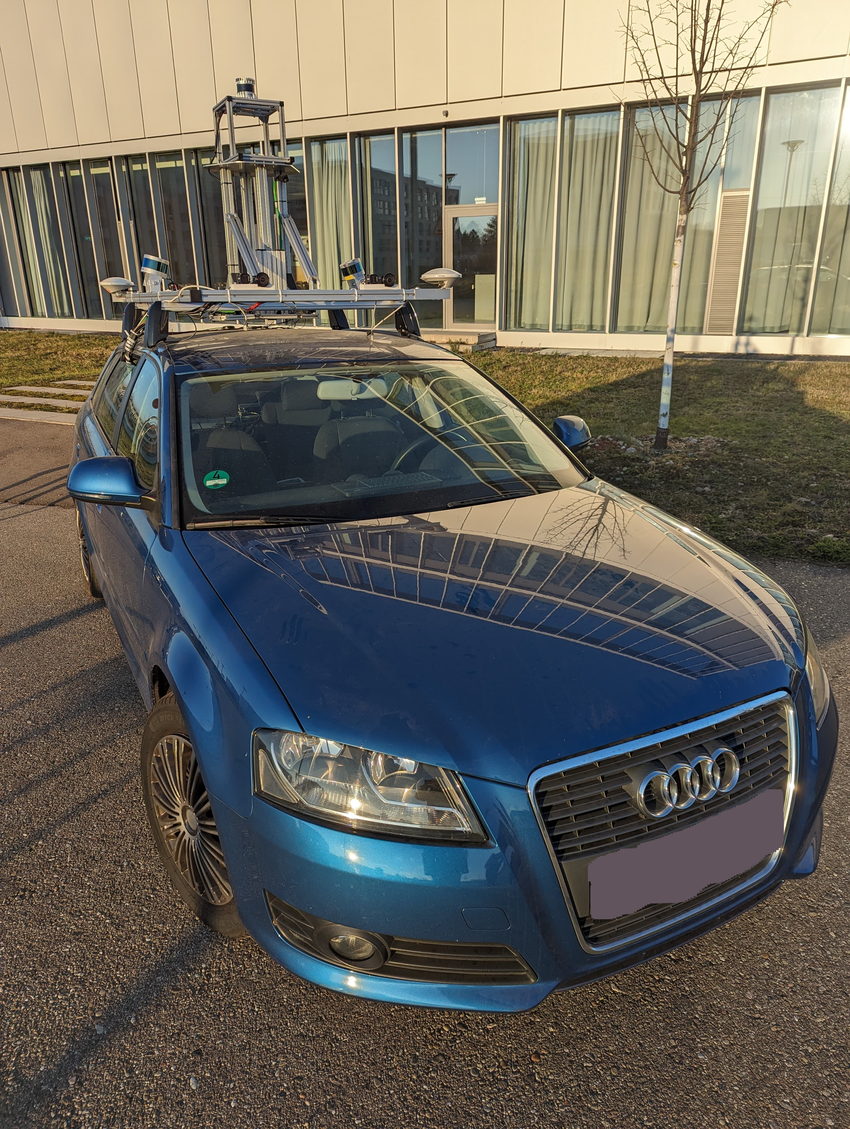}
    \caption{Our in-house autonomous driving vehicle used to demonstrate the robustness of the UP-Fuse architecture on real-world scenes.}
    \label{fig:car}
\end{figure}

\subsection{Baseline Training Details for Panoptic Waymo}
For the Panoptic Waymo dataset, we benchmark four baselines: two LiDAR-only models (EfficientLPS~\cite{sirohi2021efficientlps} and P3Former~\cite{xiao2025position}) and three multi-modal models (Panoptic-FusionNet~\cite{song2024panoptic}, LCPS~\cite{zhang2023lidar}) and IAL~\cite{pan2025images}. The LiDAR data is augmented using random yaw rotations in $[-1,1]$ radians, random scaling in $[0.9,\,1.1]$, and horizontal flipping with probability $0.5$. For camera images, we resize inputs to $256 \times 704$ and apply random scaling in $[0.5,\,2.0]$, rotations in $[-5.4^\circ,\,5.4^\circ]$, and horizontal flipping with probability $0.5$. Below, we detail the training configurations for each baseline.

{\parskip=2pt
\noindent\textbf{EfficientLPS~\cite{sirohi2021efficientlps}:}
We adopt a range-view projection of size $64 \times 2560$ pixels, which is resized to $256 \times 4096$ pixels, and train using a batch size of $8$. Following \cite{sirohi2021efficientlps}, we use stochastic gradient descent (SGD) with momentum $0.9$. The initial learning rate is $0.01$ and is decayed by a factor of $10$ at epochs $40$ and $44$. The model is trained for a total of $48$ epochs.
}

{\parskip=2pt
\noindent\textbf{P3Former~\cite{xiao2025position}:}
The 3D space is discretized into a voxel grid of $480 \times 360 \times 32$. As in \cite{xiao2025position}, we use AdamW with weight decay $0.01$, an initial learning rate of $0.005$, decaying to $0.001$ at epoch $60$. We use a batch size of $8$ and train for $80$ epochs. The loss weights follow the original implementation: classification loss ($1$), feature-segmentation losses ($1$ and $2$), and position-segmentation loss ($0.2$).
}

{\parskip=2pt
\noindent\textbf{Panoptic-FusionNet~\cite{song2024panoptic}:}
We adopt the training hyperparameters of \cite{thomas2019kpconv}, as used in the SemanticKITTI benchmark. We use a batch size of $4$ and an initial learning rate of $0.001$, optimized with SGD and a cosine annealing schedule for $80$ epochs. Loss weights are: cross-entropy ($1$), Lovász loss ($1$), center heatmap loss ($100$), and offset regression loss ($10$).
}

{\parskip=2pt
\noindent\textbf{LCPS~\cite{zhang2023lidar}:}
Following \cite{zhang2023lidar}, we voxelize the point cloud into a grid of $480 \times 360 \times 32$. The model is trained for $80$ epochs with a batch size of $4$ using the Adam optimizer. The initial learning rate of $0.004$ is reduced to $0.0004$ after $70$ epochs. Semantic supervision uses cross-entropy and Lovász losses (each weighted by $1$). BEV center heatmap regression uses an MSE loss weighted by $100$, BEV offset regression uses an L1 loss with a weight of $10$, and binary cross-entropy losses for the FOG head and region-fusion are weighted by $1$.
}

{\parskip=2pt
\noindent\textbf{IAL~\cite{pan2025images}:}
The input point cloud is voxelized into a grid of size $480 \times 360 \times 32$. The model is trained for $80$ epochs with a batch size of $4$ using the Adam optimizer. The learning rate is initialized to $4 \times 10^{-4}$, decayed to $2 \times 10^{-4}$ after $60$ epochs, and further reduced to $1 \times 10^{-4}$ after $75$ epochs. We use $128$ prior-based instance queries and $128$ no-prior instance queries.
}

\section{Qualitative Results}
\label{sec:qual_results}
\cref{fig:qual_waymo} present qualitative comparisons between our UP-Fuse network and the second-best baseline, LCPS, on the Panoptic Waymo dataset. The regions of interest are highlighted with blue and red circles, where blue indicates correct predictions, and red indicates errors made by the corresponding method.

In \cref{fig:qual_waymo} (a), we observe that LCPS misclassifies sidewalk as walkable. In this example, the grassy region geometrically resembles sidewalk, making cross-modal cues essential. UP-Fuse correctly segments the region by effectively leveraging these cross-modal cues, illustrating the benefit of our uncertainty-aware fusion. Walkable class in Panoptic Waymo corresponds to the terrain class in Panoptic nuScenes. Further, in \cref{fig:qual_waymo} (b), UP-Fuse correctly identifies a tree trunk while LCPS confuses it with a pole. This aligns with our quantitative results, where UP-Fuse achieves higher $\mathrm{PQ}^{\text{st}}$ than LCPS, confirming that our fusion strategy is particularly effective in amorphous and fine-grained \textit{stuff} regions.

In \cref{fig:qual_waymo} (c), three distant trucks are correctly detected by both methods, but LCPS misclassifies them as cars while UP-Fuse predicts the correct class. Finally, \cref{fig:qual_waymo}~(d) illustrates a night scene where LCPS fails to detect two cars due to degraded cross-modal cues, but UP-Fuse remains robust and successfully detects both instances, demonstrating improved resilience in low-light conditions.

\section{Generalization in Real-World Scenarios}
\label{sec:real}
In this experiment, we evaluate the real-world transfer capability of our UP-Fuse approach using our in-house autonomous vehicle as depicted in \cref{fig:car}. The vehicle is equipped with an Ouster 128-beam LiDAR and a front-view camera. Since the Panoptic Waymo dataset uses a 64-beam LiDAR, the higher point density of our sensor provides comparable geometric coverage. Accordingly, we use a model trained on the Panoptic Waymo dataset for this evaluation. Compared to the original Panoptic Waymo configuration, which includes five camera views, only a single front-view camera is available during inference. This setting introduces additional challenges for 3D panoptic segmentation, including domain shift, differences in sensor configurations, and reduced camera coverage. \cref{fig:real_world_qualitative} presents qualitative results from this experiment. Despite these challenges, our proposed UP-Fuse framework demonstrates promising results.

\begingroup
\hyphenpenalty=10000
\exhyphenpenalty=10000
\sloppy
\bibliographystyle{plainnat}
\bibliography{main}
\endgroup

\end{document}